\PassOptionsToPackage{pagebackref=true,breaklinks=true,colorlinks,linkcolor=black,citecolor=black,bookmarks=false}{hyperref}

\documentclass[onefignum,onetabnum]{siamart190516}

\usepackage{cite}
\usepackage{amsmath,amssymb,amsfonts}%,amsthm}
\usepackage{mathtools}
\usepackage{newtxtext,newtxmath}
\usepackage{epsfig}
\usepackage{nicefrac}
\usepackage{array}
\usepackage{color}
\usepackage{subcaption}
\usepackage{tikz}
\usepackage{tikz-cd}
\usepackage{graphicx}
\usepackage{textcomp}
\usepackage{bm}
\usepackage{algorithm}
\usepackage[noend]{algpseudocode}
\usepackage{setspace}
\usepackage{tensor}
\usepackage{xr-hyper}
\usepackage[final]{changes}
\usepackage{ifthen,calc}%
\usepackage{tikz-3dplot}
\usepackage{pgfplots}

\definechangesauthor[name={Tim Marrinan}, color=blue]{t}

\algnewcommand{\algorithmicand}{\textbf{ and }}
\algnewcommand{\algorithmicor}{\textbf{ or }}
\algnewcommand{\OR}{\algorithmicor}
\algnewcommand{\AND}{\algorithmicand}

\usepgfplotslibrary{groupplots}
\usetikzlibrary{cd,matrix,arrows,patterns,arrows,positioning}
\usetikzlibrary{shapes,calc,intersections,through}%
\usetikzlibrary{decorations.pathreplacing}
\usetikzlibrary{backgrounds}
\tikzset{subsystem/.style={%
    draw=black,rectangle, minimum size=6mm, top color=white, bottom
    color=blue!10, %thick,
    rounded corners=1mm, font=\footnotesize, }%
}%

\usepgfplotslibrary{fillbetween}
\usetikzlibrary{calc} 
\pgfplotsset{compat=newest} 
\pgfplotsset{plot coordinates/math parser=false} 
\usetikzlibrary{arrows,shapes}
\tikzstyle{every picture}+=[remember picture]
\tikzstyle{na} = [baseline=-.5ex]
\usetikzlibrary{calc}
\pgfplotsset{compat=newest} 
\pgfplotsset{plot coordinates/math parser=false} 

\def\addlegendimage{\csname pgfplots@addlegendimage\endcsname}

\DeclareMathOperator*{\argmin}{\arg\min}
\DeclareMathOperator*{\argmax}{\arg\max}
\newcommand{\Real}{\mathbb R}

\newcommand{\To}{\rightarrow}

\newlength\figureheight 
\newlength\figurewidth 
\newcommand{\etal}{\textit{et al}.\@ }

\def\*#1{\mathbf{#1}}

\usepackage{epstopdf}
\ifpdf
  \DeclareGraphicsExtensions{.eps,.pdf,.png,.jpg}
\else
  \DeclareGraphicsExtensions{.eps}
\fi

\newsiamremark{remark}{Remark}
\newsiamremark{hypothesis}{Hypothesis}
\crefname{hypothesis}{Hypothesis}{Hypotheses}
\newsiamthm{claim}{Claim}

\headers{On a minimum enclosing ball of a collection of linear subspaces}{T. Marrinan, P.-A. Absil, and N. Gillis}%, and E. Renard

\title{On a minimum enclosing ball of a\\collection of linear subspaces\thanks{Submitted to the editors 27.03.2020.
\funding{This work is supported by the Fonds de la Recherche Scientifique - FNRS and the Fonds Wetenschappelijk Onderzoek - Vlanderen (FWO) under EOS Project no O005318F-RG47. N.G. acknowledges the support of the European Research Council (ERC starting grant no 679515).}}}

\author{Tim Marrinan\thanks{\textit{Department of Mathematics and Operational Research,}  University of Mons, Mons, Belgium \newline
  (\email{timothy.marrinan@umons.ac.be}, \email{nicolas.gillis@umons.ac.be}).}
\and P.-A. Absil\thanks{\textit{ICTEAM Institute,} UCLouvain, Louvain-la-Neuve, Belgium
  (\email{pa.absil@uclouvain.be}).}%, \email{emilie.renard@uclouvain.be}).}
\and Nicolas Gillis\footnotemark[2]} %\and Emilie Renard\footnotemark[3]

\usepackage{amsopn}

\ifpdf
\hypersetup{
  pdftitle={On a minimum enclosing ball of a collection of linear subspaces},%An order-selection rule for the center of a Grassmannian minimum enclosing ball
  pdfauthor={T. Marrinan, P.-A. Absil, and N. Gillis} %and E. Renard
}
\fi

\begin{document}
\tdplotsetmaincoords{70}{110}

\maketitle

% REQUIRED
\begin{abstract}
This paper concerns the minimax center of a collection of linear subspaces.  When the subspaces are $k$-dimensional subspaces of $\Real^n$, this can be cast as finding the center of a minimum enclosing ball on a Grassmann manifold, Gr$(k,n)$.  For subspaces of different dimension, the setting becomes a disjoint union of Grassmannians rather than a single manifold, and the problem is no longer well-defined. However, natural geometric maps exist between these manifolds with a well-defined notion of distance for the images of the subspaces under the mappings. Solving the initial problem in this context leads to a candidate minimax center on each of the constituent manifolds, but does not inherently provide intuition about which candidate is the best representation of the data.  Additionally, the solutions of different rank are generally not nested so a deflationary approach will not suffice, and the problem must be solved independently on each manifold.  We propose and solve an optimization problem parametrized by the rank of the minimax center.  The solution is computed using a subgradient algorithm on the dual. By scaling the objective and penalizing the information lost by the rank-$k$ minimax center, we jointly recover an optimal dimension, $k^*$, and a central subspace, $\*U^* \in$ Gr$(k^*,n)$ at the center of the minimum enclosing ball, that best represents the data.
\end{abstract}

% REQUIRED
\begin{keywords}
  common subspace, Grassmann manifold, minimum enclosing ball, minimax center, $\ell_\infty$-center, $1$-center, circumcenter, subgradient, low-rank, order-selection
\end{keywords}

% REQUIRED
\begin{AMS}
  90C47, 14M15, 49J35  
\end{AMS}

%% Section 1
\section{Introduction}
\label{sec:intro}
Finding the minimum enclosing ball (MEB) of a finite collection of points in a metric space, or the $\ell_{\infty}$-center of mass, is a topic of broad interest in the mathematical community \cite{arnaudon2013approximating,badoiu2003smaller,renard2018grassmannian,kumar2003approximate,fischer2004smallest,yildirim2008two,nielsen2009approximating}. For Euclidean data, the problem has been well studied, and research has transitioned towards finding approximate solutions efficiently when computing the MEB exactly is impractical  \cite{badoiu2003smaller,yildirim2008two}. A breakthrough technique of B\u{a}doiu and Clarkson\cite{badoiu2003smaller} finds an optimal subset of the data, called a core-set, such that finding the exact MEB of the core-set is computationally tractable. They show that the radius of this core-set will be bounded by $(1+\epsilon)$ times the radius of the entire data set, where $\epsilon$ depends only on the number of points in the core-set~\cite{badoiu2003smaller}.  That is, the minimum enclosing ball can be approximated to any desired accuracy by increasing the number of points in the core-set, and the number of points needed for the radius of the core-set to be at most $\epsilon$ percent larger than the true radius is $\lceil \frac{2}{\epsilon} \rceil.$ This solution represents efforts to make $\ell_{\infty}$-averaging possible for complex data sets. 

The difficulty in computing the MEB of Euclidean data is due to the massive size of data sets to be averaged, however in less traditional settings other difficulties arise and contribute to the complexity of this task. Many modern problems are formulated on manifolds instead of Euclidean space in situations where the manifold geometry better represents the natural structure of the data model~\cite{chellapa,rentmeesters2010efficient,marrinan2016}.  Afsari provided existence and uniqueness conditions for Riemannian $\ell_p$ centers of mass~\cite{afsari2011riemannian}, and with this type of structure in mind, Arnaudon and Nielsen~\cite{arnaudon2013approximating} adapted the efficient MEB algorithm of B\u{a}doiu and Clarkson to Riemannian manifolds. For linear subspace data, a subclass of data addressed by~\cite{arnaudon2013approximating}, this work was further generalized by Renard, Gallivan, and Absil~\cite{renard2018grassmannian,renard2019minimax}.  They created a technique that applies to points lying on a disjoint union of Grassmann manifolds, that is, a collection of $p_i$-dimensional subspaces of $\Real^n$ where $p_i$ is not necessarily equal for all $i$. Although the data comes from a collection of manifolds, the MEB must be computed on one individual Grassmannian and the choice of which is not obvious. Determining which Grassmannian provides the best center for a collection of subspaces is one of the tasks of this manuscript, and we provide a geometrically motivated criteria for automatically selecting this manifold.

%% Common subspace illustration
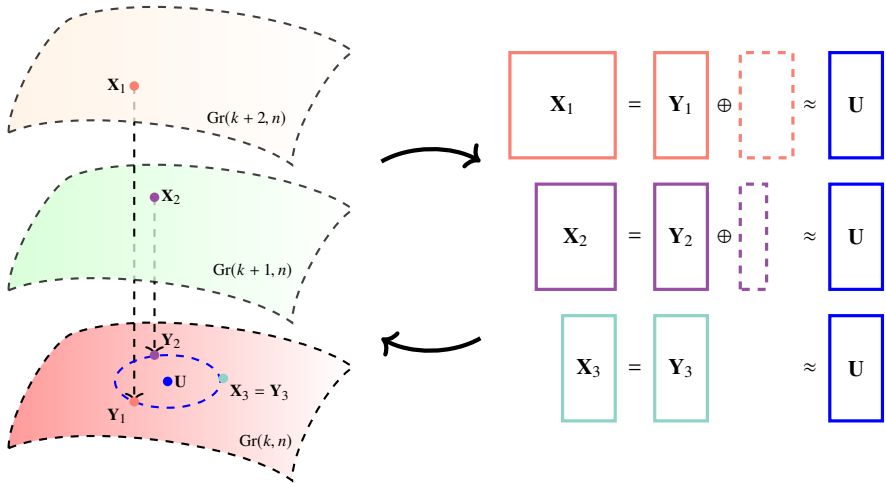
\begin{figure*}
\centering
%% Figure 1

\definecolor{mycolor1}{rgb}{0.98431,0.50196,0.44706}%
\definecolor{mycolor2}{rgb}{0.59608,0.30588,0.63922}%
\definecolor{mycolor3}{rgb}{0.99216,0.70588,0.38431}%
\definecolor{mycolor4}{rgb}{0.55294,0.82745,0.78039}%

\begin{tikzpicture}[thick,scale=0.7, every node/.style={transform shape}]

%% Manifolds

	% Bottom manifold, Gr(k,n)
	\draw[dashed,looseness=.6, shade, left color=red!40!white] (-3,-5,2)
  to[bend left] (2,-6,1)
  to[bend left] coordinate (mp) (2,-5,-2)
  to[bend right] (-3,-4,-1)
  to[bend right] coordinate (mm) (-3,-5,2)
  -- cycle;
	\node[] at (1.5,-5.25,1){\small Gr$(k,n)$};

	% Projection lines
	\draw [thick, dashed, ->] (-1,-1.5,1) to (-1,-4.5,1);% middle manifold to Y1
	\draw [thick, dashed, ->] (-1,-1,0) to (-1,-4,0); % X2 to Y2
	
	% Middle manifold Gr(k+1,n)
	\draw[dashed,looseness=.6, shade, left color=green!20!white, opacity=0.75] (-3,-2,2) 
  to[bend left] (2,-3,1)
  to[bend left] coordinate (mp) (2,-2,-2)
  to[bend right] (-3,-1,-1)
  to[bend right] coordinate (mm) (-3,-2,2)
  -- cycle;
	\node[] at (1.25,-2,1){\small Gr$(k+1,n)$};
	
	% Projection lines
	\draw [thick, dashed] (-1,1.5,1) to (-1,-1.5,1); %X1 to middle manifold
	
	% Top manifold, Gr(k+2,n)
	\draw[dashed,looseness=.6, shade, left color=mycolor3!20!white,opacity=0.75] (-3,1,2)
  to[bend left] (2,0,1)
  to[bend left] coordinate (mp) (2,1,-2)
  to[bend right] (-3,2,-1)
  to[bend right] coordinate (mm) (-3,1,2)
  -- cycle;
	\node[] at (0.75,0.5,0){\small Gr$(k+2,n)$};

%% Points
	\draw[blue,thick,dashed] (-0.75,-4.5,0) ellipse (1cm and .5cm); % Minimum enclosing ball
	
	\filldraw[mycolor1,fill=mycolor1](-1,1.5,1)circle (2pt) node[left] {};
	\node[left] at (-1,1.5,1){\small $\*X_1$};	
	
	\filldraw[mycolor2,fill=mycolor2](-1,-1,0)circle (2pt) node[right] {};
	\node[right] at (-1,-1,0){\small $\*X_2 \ $};

	\filldraw[mycolor1,fill=mycolor1](-1,-4.5,1)circle (2pt) node[below left] {};
	\node[below left] at (-1,-4.5,1) {\small $\*Y_1$};

	\filldraw[mycolor2,fill=mycolor2](-1,-4,0)circle (2pt) node[above right] {};
	\node[above right] at (-1,-4,0) {\small $\*Y_2$};

	\filldraw[mycolor4,fill=mycolor4](0.5,-4.25,.5)circle (2pt) node[above right] {};
	\node[below right] at (0.5,-4.25,.5) {\small $\*X_3 = \*Y_3$};

	\filldraw[blue,fill=blue](-0.75,-4.5,0)circle (2pt) node[left] (Z) {};
	\node[right] at (-0.75,-4.5,0){\small $\*U$};
	
%% Flow chart arrows
	\draw [ultra thick, shorten >=0.25cm,shorten <=.25cm, ->] (3,-0.5,0) to[bend left] (5.5,-0.5,0);
	\draw [ultra thick, shorten >=0.25cm,shorten <=.25cm, <-] (3,-3.5,0) to[bend right] (5.5,-3.5,0);		

%% Subspace bases
	% X1
  \draw[mycolor1,very thick] (5.75,-0.25,0) rectangle (7.75,1.75,0);
	\node[] at (6.75,0.75,0){\large $\*X_1$};
	\node[] at (8.125,0.75,0){\large $=$};
	% Y1
	\draw[mycolor1,very thick] (8.5,-0.25,0) rectangle (9.5,1.75,0);
	\node[] at (9,0.75,0){\large $\*Y_1$};
	\node[] at (9.825,0.75,0){\large $\oplus$};
	% Orthogonal complement
	\draw[mycolor1,dashed,very thick] (10.125,-0.25,0) rectangle (11.125,1.75,0);
	% U
	\node[] at (11.45,0.75,0){\large $\approx$};
	\draw[blue,very thick] (11.825,-0.25,0) rectangle (12.825,1.75,0);
	\node[] at (12.325,0.75,0){\large $\*U$};
	
	% X2
	\draw[mycolor2,very thick] (6.25,-2.75,0) rectangle (7.75,-0.75,0);
	\node[] at (7,-1.75,0){\large $\*X_2$};
	\node[] at (8.125,-1.75,0){\large $=$};
	% Y2
	\draw[mycolor2,very thick] (8.5,-2.75,0) rectangle (9.5,-0.75,0);
	\node[] at (9,-1.75,0){\large $\*Y_2$};
	\node[] at (9.825,-1.75,0){\large $\oplus$};
	% Orthogonal complement
	\draw[mycolor2,dashed,very thick] (10.125,-2.75,0) rectangle (10.625,-.75,0);
	% U
	\node[] at (11.45,-1.75,0){\large $\approx$};
	\draw[blue,very thick] (11.825,-2.75,0) rectangle (12.825,-0.75,0);
	\node[] at (12.325,-1.75,0){\large $\*U$};	
	
	% X3
	\draw[mycolor4,very thick] (6.75,-5.25,0) rectangle (7.75,-3.25,0);
	\node[] at (7.25,-4.25,0){\large $\*X_3$};
	\node[] at (8.125,-4.25,0){\large $=$};
	% Y3
	\draw[mycolor4,very thick] (8.5,-5.25,0) rectangle (9.5,-3.25,0);
	\node[] at (9,-4.25,0){\large $\*Y_3$};
	% U
	\node[] at (11.45,-4.25,0){\large $\approx$};
	\draw[blue,very thick] (11.825,-5.25,0) rectangle (12.825,-3.25,0);
	\node[] at (12.325,-4.25,0){\large $\*U$};

\end{tikzpicture}
\caption{\label{fig:common_info}One way to interpret the center of the Grassmannian minimum enclosing ball is as a basis for the common information in a collection of subspaces. If the subspaces share information in $k$ dimensions, then each subspace in the collection contains a $k$-dimensional subspace, $\*Y_i \subseteq \*X_i$ for $i=1,2,3$, that represents this similarity. If $\*U$ is the minimax center of these points on Gr$(k,n)$, then $\*U$ is the best $k$-dimensional approximation to the original subspaces $\*X_1,\*X_2,$ and $\*X_3$ in the sense that it minimizes the maximum distance to $\{\*Y_i\}_{i=1}^3.$}
\end{figure*}

With subspace data, it is natural to think of the center of the Grassmannian minimum enclosing ball (GMEB) as the common information in the data set. To see this, consider the illustration in Figure~\ref{fig:common_info}. Suppose that $\{\*X_i\}_{i=1}^3$ are linear subspaces of $\Real^n$ with dim$(\*X_1) = k+2,$ dim$(\*X_2) = k+1,$ and dim$(\*X_3) = k.$ In Figure~\ref{fig:common_info}, these subspaces are indicated by the colored rectangles on the right. The subspaces can be identified with points on different Grassmannians, which are pictured by the corresponding colored points on the lefthand side of the figure. If these three spaces intersect in some $k$-dimensional subspace $\*U,$ then $\*U$ is certainly one of the best $k$-dimensional approximations of the collection.  Alternatively, if the spaces do not intersect, we can look for a best $k$-dimensional approximation, $\*U,$ to these spaces by minimizing some measure of dissimilarity to the collection. One formulation is to find $\*U$ such that it minimizes the maximum dissimilarity with the elements of $\{\*X_i\}_{i=1}^3.$ Once this $\*U$ is identified, there is an implicitly defined $k$-dimensional subspace for each element in the set, $\*Y_i \subseteq \*X_i$ for $i=1,2,3,$ with the property that $\*U$ is the best $k$-dimensional approximation for $\{\*Y_i\}_{i=1}^3.$ This property can be seen in Figure~\ref{fig:common_info} where $\*U$ is the center of the minimum enclosing ball of the points $\*Y_i$ associated with each $\*X_i.$

Common subspace extraction can be found in subspace clustering~\cite{abdolali2019scalable}, domain adaptation, and subspace alignment. These tools can be used in a plethora of tasks in pattern recognition including subspace tracking~\cite{srivastava2004bayesian}, face recognition~\cite{chang2012feature,chakraborty2015recursive}, video action recognition~\cite{o2012scalable,chakraborty2015recursive}, infected patient diagnosis~\cite{ma2018self}, adaptive sorting~\cite{jurrus2016adaptive}, model reduction~\cite{franz2014interpolation}, and many more.  Common subspace extraction is frequently done by finding the $\ell_2$- or $\ell_1$-center in cases where outliers are present in the data collection, but if the data are drawn from a uniform distribution whose support is a ball, the $\ell_{\infty}$-center gives the maximum likelihood estimator for the center of the support and thus may be preferred when all the subspaces are assumed to be valid~\cite{afsari2011riemannian}. Furthermore, techniques have been developed to prune outliers from data sets using the $\ell_{\infty}$-norm, with theoretical guarantees in some circumstances~\cite{sim2006removing}.

In this paper, we present a novel technique to accurately estimate the GMEB for a collection of linear subspaces of possibly differing dimension, and a geometrically inspired order-selection rule to identify the Grassmannian that best represents the shared information in the data. Choosing the ideal manifold on which to perform the $\ell_{\infty}$-averaging is inherently related to finding a common subspace of optimal rank, and thus the numerical experiments explore the relationships between different rank-adaptive subspace averaging methods.

The main contributions of the paper are summarized as follows. We propose

\begin{itemize}
\item a subgradient approach to solve the dual of the GMEB problem for subspaces of differing dimensions. A duality gap of zero certifies the solution as optimal.
\item an unsupervised order-selection rule for the dimension of the center of the GMEB.
\item a warm-start initialization for the subgradient algorithm that reduces the number of iterations needed for the subgradient algorithm to converge.
\item a hybrid method for order-selection which modifies the existing rule of~\cite{santamaria2016order} for use with the center of the GMEB.
\item a synthetic data model that allows us to measure the accuracy of an estimate for the center of the GMEB, and demonstrate the effectiveness of the proposed technique using data generated with this model.
\end{itemize} 
Finally, we compare the proposed order-selection rules to existing methods for automatic order selection in subspace averaging with numerical experiments.

%% SECTION 2
\section{Problem formulation: Grassmannian minimum enclosing ball}
\label{sec:p2s}
In this section we provide the mathematical background necessary to formulate the GMEB problem for subspaces of differing dimension. We define maps that associate a subset of points on a single manifold with each subspace from the collection, and we describe the point-to-set distance that measures the dissimilarity of these sets. Finally, we explicitly state the minimax optimization problem that defines this GMEB.
 
Denote by Gr$(k,n)$ the Grassmann manifold of $k$-dimensional subspaces in $\Real^n$. If $A$ is an $n \times k$ matrix with full column rank, the column space of $A$, col$(A)$,  defines a subspace that can be identified with a point $\*A \in \textrm{Gr}(k,n)$. To simplify notation we assume without loss of generality that the chosen representative for a point $\*A \in \textrm{Gr}(k,n)$ is an orthonormal basis, $A \in \Real^{n \times k}$ with $A^TA = I$. Let $\textrm{O}(k)$ denote the set of $k \times k$ orthogonal matrices. If $Q_k \in \textrm{O}(k)$ then $\textrm{col}(A Q_k) = \textrm{col}(A) = \*A ,$ and we can see that a point on this Grassmannian can be represented by any real $n \times k$ matrix that spans the same subspace. For any two points, $\*A, \*B \in \textrm{Gr}(k,n),$ there exists a set of $k$ principal angles,  $0 \leq \theta_1(\*A,\*B) \leq \cdots \leq \theta_k(\*A,\*B) \leq \nicefrac{\pi}{2},$ defined recursively as
\begin{equation}
\label{eq:angles}
\begin{aligned}
\theta_1(\*A,\*B) := &\underset{\*a_1 \in \*A, \*b_1 \in \*B}{\min} \cos^{-1} \left(\frac{\*a_1^T\*b_1^{}}{\|\*a_1\|_2 \|\*b_1\|_2} \right) , \textrm{ and for } i = 2, \ldots, k\\
\theta_i(\*A,\*B) := &\underset{\*a_i \in \*A, \*b_i \in \*B}{\min} \cos^{-1} \left(\frac{\*a_i^T\*b_i}{\|\*a_i\|_2 \|\*b_i\|_2} \right)\\
& \ \textrm{s.t. } \*a_j^T \*a_i^{} = 0 \textrm{ for } j<i\\
& \ \phantom{\textrm{s.t. }} \*b_j^T \*b_i^{} = 0 \textrm{ for } j<i.
\end{aligned}
\end{equation}
The vectors that form these angles, $\{\*a_1, \ldots, \*a_k\}$ and $\{\*b_1, \ldots, \*b_k\},$ are called the left and right principal vectors, respectively, and form orthogonal bases for the spaces $\*A$ and $\*B$. The principal angles and principal vectors can be computed via the singular value decomposition (SVD)~\cite{golub}. Let $A^T B = V \Sigma W^T$ be a thin SVD with the singular values sorted in nonincreasing order, so that 
\begin{equation}
\label{eq:svd}
\begin{aligned}
V \in \Real^{k\times k} &\textrm{ with } V^TV = I, \\
\Sigma \in \Real^{k \times k} &\textrm{ with } \Sigma = \textrm{diag}(\cos(\bm{\theta}(\*A,\*B))), \textrm{ and}\\
W \in \Real^{k\times k} &\textrm{ with } W^T W = I.
\end{aligned}
\end{equation} 
Then $\theta_i(\*A,\*B) = \cos^{-1}(\Sigma_{ii})$ is the $i$th principal angle separating $\*A$ and $\*B$, with associated left and right principal vectors $\*a_i = A\*v_i$ and $\*b_i = B\*w_i$ for $i = 1, \ldots, k$.

Let $d: \textrm{Gr}(k,n) \times \textrm{Gr}(k,n) \To \Real$ be a Grassmannian metric. If for all $\*A, \*B \in \textrm{Gr}(k,n)$ and for all $Q_n \in \textrm{O}(n)$ the left action of $Q_n$ on $A$ and $B$ by multiplication does not change the value of the metric, that is, $d(\*A,\*B) = d(\*{Q_nA},\*{Q_nB}),$ then $d$ is said to be orthogonally invariant. Orthogonally invariant metrics depend only on the relative position of $\*A$ and $\*B,$ so as a result of~\cite[Thm.~3]{wong1967differential}, $d$ can be written as a function of the vector of principal angles separating $\*A$ and $\*B,$ $\bm{\theta}(\*A,\*B) \in \Real^{k}$. Additionally, for $Gr(k,n)$ with either $k\neq2$ or $n\neq2$ there is an essentially unique invariant Riemannian metric (up to scaling) which yields $d(\*A,\*B) = \| \bm{\theta}(\*A,\*B) \|_2,$ and is frequently referred to as the geodesic distance based on arc length~\cite{wong1967differential}. For an orthogonally invariant metric $d(\cdot,\cdot),$ the generalized Grassmann mean of $\left\{\*X_i\right\}_{i=1}^M \in \textrm{Gr}(k,n)$ is defined as
\begin{equation}
\label{eq:generalized_mean}
\*U^* =  \underset{\*U \in \textrm{Gr}(k,n)}{\argmin} \left(\sum_{i=1}^M d(\*U,\*X_i)^p\right)^{\nicefrac{1}{p}}.
\end{equation}
When $p=2$ the solution is the Grassmannian center of mass, or the Karcher mean~\cite{karcher}.

This manuscript is concerned with computing the generalized Grassmann mean when $p \to \infty.$ However, rather using a Grassmannian metric we measure dissimilarity by the squared chordal distance, $d(\*A,\*B) = \|\sin(\bm{\theta}(\*A,\*B))\|_2^2.$ A common interpretation of $\ell_{\infty}$-norm minimization is that it minimizes the maximum value.  In this context we wish to solve,
\begin{equation}
\label{eq:minmax}
\*U^* =   \underset{\*U \in \textrm{Gr}(k,n)}{\argmin} \lim_{p \to \infty} \left(\sum_{i=1}^M d(\*U,\*X_i)^{p}\right)^{\nicefrac{1}{p}} = \underset{\*U \in \textrm{Gr}(k,n)}{\argmin} \max_{i = 1, \ldots, M} d(\*U,\*X_i),
\end{equation}
for a collection of Grassmannian points, $\left\{\*X_i\right\}_{i=1}^M$.  The solution, $\*U^{*},$ can be referred to as the minimax center, and is the center of the minimum enclosing ball of the collection on Gr$(k,n).$ 

Alternatively, let $\mathcal{D} = \left\{\*X_i \right\}_{i=1}^M$ be a finite collection of subspaces of $\Real^n$ with possibly different dimensions, so that dim$(\*X_i) = p_i.$ For the set of positive integers $\mathcal{P} = \{\textrm{dim}(\*X_i) : \*X_i\in\mathcal{D} \}$ we can consider $\mathcal{D}$ as a collection of points lying on the disjoint union of Grassmann manifolds, $\*X_i \in \coprod_{p \in \mathcal{P}}{\textrm{Gr}(p,n)}.$ In this scenario Equation~\eqref{eq:minmax} is not well-defined without further formalism. To account for the difference in subspace dimensions, we adopt the convention of~\cite{ye_lim} by redefining $d(\*U,\*X_i)$ as the minimum distance between $\*U$ and a subset of points on $\textrm{Gr}(k,n),$ appropriately defined for each $\*X_i \in \mathcal{D}$. Each subspace is associated with one of two types of subset, which are defined by
\begin{equation}
\label{eq:schubs}
\begin{aligned}
 \Omega_{+}(\*X_i) \doteq \left\{\*Y \in \textrm{Gr}(k,n) : \*X_i\subseteq \*Y \right\} &\textrm{ for }  p_i < k, \textrm{ and}   \\
\Omega_{-}(\*X_i) \doteq \left\{\*Y \in \textrm{Gr}(k,n) : \*Y \subseteq \*X_i \right\} & \textrm{ for } p_i \geq k.
\end{aligned}
\end{equation}
We use $\Omega_{*}(\*X_i)$ when referring to either type generically. For $\*X_i$ such that $p_i < k$, $\Omega_{+}(\*X_i)$ is the set of all points of Gr$(k,n)$ containing $\*X_i.$ Alternatively when $\*X_i$ is a $p_i$-plane with $p_i > k,$ $\Omega_{-}(\*X_i)$ is all $k$-dimensional subspaces contained in $\*X_i,$ and when $p_i = k$ the subset of points is just the singleton, $\*X_i$.

\begin{figure*}[t]
\centering
%% Figure 2
\begin{tikzpicture}[tdplot_main_coords]
	
	% Left manifold, Gr(k+1,n)
  \draw[dashed,looseness=.6, shade, left color=green!60!black] (0,-6,2)
  to[bend left] (0,-3,2)
  to[bend left] coordinate (mp) (-4,-3,2)
  to[bend right] (-4,-6,2)
  to[bend right] coordinate (mm) (0,-6,2)
  -- cycle;
	\node[] at (0,-5,2){\small Gr$(k+1,n)$};
	\filldraw (-2,-3.5,2.5)circle (2pt) node[left] (X) {$\*X_1$};
	
	% Middle manifold, Gr(k,n)
	\draw[dashed,looseness=.6, shade, left color=red!40!white] (3,-3,1)
  to[bend left] (3,3,1)
  to[bend left] coordinate (mp) (-5,3,1)
  to[bend right] (-5,-3,1)
  to[bend right] coordinate (mm) (3,-3,1)
  -- cycle;
	\node[] at (2.5,-0.5,1){\small Gr$(k,n)$};
	
	% Right manifold, Gr(k-1,n)
	\draw[dashed,looseness=.6, shade, left color=blue!60!black] (0,3,0)
  to[bend left] (0,6,0)
  to[bend left] coordinate (mp) (-4,6,0)
  to[bend right] (-4,3,0)
  to[bend right] coordinate (mm) (0,3,0)
  -- cycle;
	\node[] at (0,4,0){\small Gr$(k-1,n)$};
	\filldraw (-2,4.5,0.5)circle (2pt) node[right] (Y) {\small $\*X_2$};
	
	% Schubert variety of X2
	\draw[ultra thick, blue!60!black, rounded corners, name path=line 1] (-1,-0.25,1)
  to[bend left] (-.5,1.5,1)
  to[bend right] coordinate (Ymap1) (-3,1.5,1)
  to[bend right] (-3,-0.25,1.3)
  to[bend right] coordinate (Ymap2) (-1,-0.25,1)
  -- cycle;	
	\node[] at (-2.5,0.5,1.25){\small $\Omega_{+}(\*X_2)$};	
  
	% Schubert variety of X1
	\draw[ultra thick, green!60!black, rounded corners, name path=line 2]  (0,-2.5,1)
  to[bend left] (0,-0.75,1)
  to[bend right] coordinate (Ymap3) (-2,-0.75,1)
  to[bend right] (-2,-2.5,1.3)
  to[bend right] coordinate (Ymap4) (0,-2.5,1)
  -- cycle;		
	\node[] at (-1.5,-2,1.25){\small $\Omega_{-}(\*X_1)$};
	
	% Points
	\filldraw(0,0.75,1)circle (2pt) node[below right]{$\*U$};
	\filldraw(-1.25,0.75,1)circle (2pt) node[right] (Z) {\small $\ \*Y_2$};
	\filldraw(-1,-0.5,1)circle (2pt) node[left] (W) {\small $\*Y_1 \ $};
	
	% Shortest paths to U
  \draw[dotted, thick] (0,0.75,1) to[bend left] (-1.25,0.75,1);
	\draw[dotted, thick] (0,0.75,1) to[out=130,in=-5] (-1,-0.5,1);
	
	% Maps between manifolds
	\draw [ultra thick,shorten >=0.25cm,shorten <=.25cm, ->] (-2,4.5,0.5) to[bend right] (-3,1.5,1); 
	\draw [ultra thick, shorten >=0.25cm,shorten <=.25cm, ->] (-2,-3.5,2.5) to[out=10,in=90] (-2.75,-2,1); 
\end{tikzpicture}
\caption{\label{fig:schubs}Illustration of the minimum point-to-set distance on $\textrm{Gr}(k,n)$ between $\*U$ and the sets $\Omega_{-}(\*X_1)$ and $\Omega_{+}(\*X_2)$ associated with points on $\textrm{Gr}(k+1,n)$ and $\textrm{Gr}(k-1,n)$, respectively. The points that realize the minimum distance are $\*Y_1 \in \Omega_{-}(\*X_1)$ and $\*Y_2 \in \Omega_{+}(\*X_2)$.}
\end{figure*}
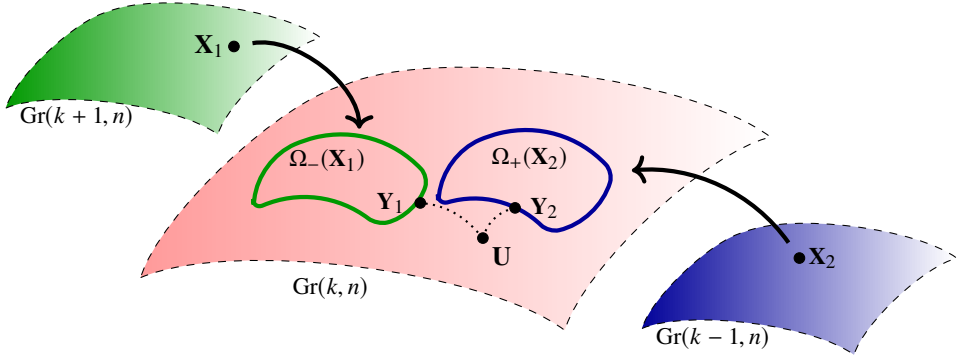

Finally,  we overload the notation for distance so that 
\begin{equation}
\label{eq:schub_dist}
d_{\textrm{Gr}(k,n)}(\*U,\*X_i) \doteq d_{\textrm{Gr}(k,n)}(\*U,\Omega_{*}(\*X_i)) = \min \{ d(\*U,\*Y_i) : \*Y_i \in \Omega_{*}(\*X_i)\} 
\end{equation}
when the distance is being measured on $\textrm{Gr}(k,n)$ and the data comes from Grassmann manifolds of possibly differing dimension. This is the proposed distance of~\cite{ye_lim}, which is well-defined on any single fixed Grassmannian. Figure~\ref{fig:schubs} shows an illustration of this distance as the length of the shortest path between a point, $\*U$, and the set of points, $\Omega_{*}(\*X_i)$.

%% Schubert variety illustration

The minimum in Equation~\eqref{eq:schub_dist} always exists because $\Omega_{*}(\*X_i)$ is a closed subset of the Grassmannian, and the points satisfying $\*Y_i = \argmin_{\*Y \in \Omega_{*}(\*X_i)} d(\*U,\*Y)$ are independent of the choice of orthogonally invariant distance measure~\cite{schwickerath2014linear}. Let $U^TX_i = V \Sigma W^T$ be a thin SVD as in Equation~\eqref{eq:svd}. One point that achieves the minimum distance is the columnspace of the matrix defined by
\begin{equation}
\label{eq:y_def}
Y_i \doteq 
\begin{dcases} 
      \left[X_i\*w_1, \ldots,X_i\*w_k\right] & \textrm{for } p_i\geq k; \\[4pt]
      \left[X_i\*w_1, \ldots,X_i\*w_{p_i}, U\*v_{p_i+1},\ldots, U\*v_k\right] & \textrm{otherwise.}
\end{dcases}
\end{equation}

This formalism implies that distances can be written as a function of exactly $k$ principal angles regardless of the dimension of $\*X_i$, and conveniently the definition agrees with many pseudo-metrics commonly used in the literature that measure similarity as a function of the (possibly less than $k$) principal angles between subspaces of different dimension. It should be clear, however, that this is not a metric because the distance between $\*A$ and $\*B$ will be zero if $\*A$ is a proper subspace of $\*B$, despite being non-identical. 

The minimum point-to-set distance using the squared chordal distance is
\begin{equation}
\begin{aligned}
\label{eq:p2s_dist}
d_{\textrm{Gr}(k,n)}(\*U,\*X_i) &=\|\sin(\bm{\theta}(\*U,\*Y_i))\|_2^2 \\
&=  \frac{1}{2}\|U^{}_k U^{T}_k - Y_i^{} Y_i^T \|_F^2 \\
& = k - \textrm{Tr}(U^{T}Y_i^{} Y_i^{T}U^{}) \\
&= \min \{k,p_i\} - \textrm{Tr}(U^TX_i^{}X_i^TU),
\end{aligned}
\end{equation}
where $\bm{\theta}(\*U,\*Y_i) \in \mathbb{R}^k$ is the vector of principal angles between $\*U$ and the point $\*Y_i \in \Omega_{*}(\*X_i)$ that attains the minimum. The final equality in Equation~\eqref{eq:p2s_dist} can be seen from the definition of $\*Y_i$ in Equation~\eqref{eq:y_def} and will be demonstrated in Equation~\eqref{eq:chordal}. Note that  it is not necessary to know $\*Y_i$ in order to compute $d_{\textrm{Gr}(k,n)}(\*U,\*X_i).$ With this definition and choice distance measurement, the problem in~\eqref{eq:minmax} is well-defined when written as
\begin{equation}
\label{eq:formal_prob}
\*U^* = \underset{\*U \in \textrm{Gr}(k,n)}{\argmin} \max_{i = 1, \ldots, M} d_{\textrm{Gr}(k,n)}(\*U,\*X_i).
\end{equation}

Using the notion of distance from Equation~\eqref{eq:schub_dist}, an algorithm was proposed by \cite{renard2018grassmannian} to solve Problem~\eqref{eq:formal_prob} for a given value of $k$. Since the data is not of uniform dimension, it is one of our goals to find the solution across all possible values of $k$ that best represents the common subspace in the data. In Section~\ref{sec:ord_select} we propose an order-selection rule for comparing solutions of different dimension, however we must first be able to find the solutions of different dimension efficiently. In general  $\*U^*(k) \in \textrm{Gr}(k,n)$ is not contained in $\*U^*({k+1}) \in \textrm{Gr}(k+1,n),$ so it is not possible to construct the respective solutions iteratively via deflation.  Instead the problem needs to be solved independently for each value of $k$.

%% Section 3
\section{Dual formulation}
\label{sec:existing}
Problem~\eqref{eq:formal_prob} is nonconvex and challenging to optimize directly. Therefore, in this section we formulate its dual function which can be solved efficiently. The dual variables also provide a primal-feasible solution, which can be tested for optimality.

Using Equation~\eqref{eq:p2s_dist}, Problem~\eqref{eq:formal_prob} can be written as one with matrix arguments that can be identified with the Grassmannian points they represent. That is, 
\begin{equation}
\label{eq:matrix_eq}
\begin{aligned}
U^* 
= & \ \underset{U \in \Real^{n \times k}}{\argmin} \max_{i = 1, \ldots, M} \left(\min \{k,p_i\} - \textrm{Tr}(U^TX_i^{}X_i^TU) \right)\\
& \ \textrm{s.t. } U^TU = I,
\end{aligned}
\end{equation}
where  $U$ is an orthonormal basis for $\*U,$ $X_i$ is an orthonormal basis for $\*X_i,$ and $p_i = \dim(\*X_i).$ The solution to \eqref{eq:formal_prob} is then the column space of the solution to \eqref{eq:matrix_eq}, $\*U^* = \textrm{col}(U^*)$. For ease of notation we will treat the dual problem as a minimization, so we reformulate the primal as,
\begin{equation}
\begin{aligned}
U^* 
= & \ \underset{U \in \Real^{n \times k}}{\argmax} \min_{i = 1, \ldots, M} - \left(\min \{k,p_i\} - \textrm{Tr}(U^TX_i^{}X_i^TU) \right)\\
& \ \textrm{s.t. } U^TU = I.
\end{aligned}
\end{equation}
Adding an auxiliary variable $\tau$, the quadratic cost function to be minimized is replaced by a smooth linear objective that is maximized with respect to quadratic inequality constraints,
\begin{subequations}
\begin{align}
U^* 
= & \ \underset{U \in \Real^{n \times k}, \ \tau \in \Real}{\argmax} \tau \\
& \ \textrm{s.t. } -\tau -  \min \{k,p_i\} + \textrm{Tr}(U^TX_i^{}X_i^TU)  \geq 0 \textrm{ for } i = 1, \ldots, M, \label{eq:const1}\\
& \ \hphantom{s.t. }  U^TU = I. \label{eq:const2}
\end{align}
\end{subequations}

Let $\bm{\lambda} = [\lambda_1, \ldots, \lambda_M]^{T}$ be a vector of Lagrange multipliers associated with the inequality constraints in~\eqref{eq:const1}. Dualizing only the inequality constraints leads to the Lagrangian%
\begin{equation}
\begin{aligned}
\mathcal{L}(U,\tau,\bm{\lambda}) &=  \tau  + \sum_{i=1}^M \lambda_i \left(-\tau -\min \{k,p_i\} + \textrm{Tr}(U^TX_i^{}X_i^TU) \right),   \label{eq:lagrange}
\end{aligned}
\end{equation}
such that $U^TU=I$ and $\lambda_i \geq 0$ for $i = 1,\ldots, M,$ with first-order optimality conditions
\begin{subequations}
\begin{align}
\sum_{i=1}^M \lambda_i &= 1 && \left(\nabla_{\tau} \mathcal{L}(U,\tau,\bm{\lambda}) = 0 \right),\label{eq:kkt2}\\
-\tau - \min \{k,p_i\} + \textrm{Tr}(U^TX_i^{}X_i^TU)  &\geq 0 \ \textrm{for } i = 1, \ldots, M &&  \left(\nabla_{\bm{\lambda}} \mathcal{L}(U, \tau,\bm{\lambda}) \geq 0 \right),\label{eq:kkt5}\\
\lambda_i\big(\tau + \min \{k,p_i\} - \textrm{Tr}(U^TX_i^{}X_i^TU) \big) &= 0  \ \textrm{for } i = 1, \ldots, M && (\textrm{complementarity}), \label{eq:kkt6} \\
\lambda_i &\geq 0 \ \textrm{for } i = 1, \ldots, M && (\textrm{nonnegativity}) \label{eq:kkt1}.
\end{align}
\end{subequations}

The dual of Equation~\eqref{eq:lagrange} is found by maximizing $\mathcal{L}$ over $U$ and $\tau$,
\begin{equation}
\begin{aligned}
f(\bm{\lambda}) &= \sup_{\tau}\big( \tau  - \sum_{i=1}^M \lambda_i \tau \big) + \sup_{U^TU = I} \left( \sum_{i=1}^M -\lambda_i^{}\big(\min \{k,p_i\} - \textrm{Tr}(U^TX_i^{}X_i^TU) \big)\right).   
\end{aligned}
\end{equation}
The maximum over $\tau$ yields $f(\bm{\lambda}) = \infty$ unless $\|\bm{\lambda}\|_1 = 1,$ in which case the first term is zero, and the dual can be written as
\begin{equation}
\label{eq:dual_sup}
\begin{aligned}
f(\bm{\lambda}) &= -\sum_{i=1}^M \lambda_i^{} \min \{k,p_i\}  + \sup_{U^TU = I}  \textrm{Tr}(U^T(\sum_{i=1}^M \lambda_iX_i^{}X_i^T ) U)
\end{aligned}
\end{equation}
The set of $n \times k$ matrices with orthonormal columns is closed, thus the supremum is achieved by an element of the set. For a given $\bm{\lambda} \in \Real^{M},$ let $\big(\sum_{i=1}^M \lambda_i^{}X_i^{} X_i^{T}\big) V=  VD$ be an orthogonal eigenvector decomposition where the eigenvalues are ordered in decreasing magnitude, $D_{11} \geq D_{22} \geq \cdots \geq D_{nn}.$ The matrix whose columns are the $k$ dominant eigenvectors, 
\begin{equation}
\label{eq:weightedEVD}
U_{\bm{\lambda}} \doteq [\*v_1, \ldots, \*v_k],
\end{equation}
satisfies $U_{\bm{\lambda}}^TU_{\bm{\lambda}}^{} = I$ and maximizes the term $\textrm{Tr}(U^T(\sum_{i=1}^M \lambda_iX_i^{}X_i^T ) U),$ so we can write
\begin{equation}
\label{eq:dual_max}
\begin{aligned}
f(\bm{\lambda}) &= -\sum_{i=1}^M \lambda_i^{} \min \{k,p_i\}  + \textrm{Tr}(U_{\bm{\lambda}}^T(\sum_{i=1}^M \lambda_iX_i^{}X_i^T ) U_{\bm{\lambda}}^{}).
\end{aligned}
\end{equation} 
Finally, we wish solve this optimization problem,
\begin{equation}
\label{eq:dualProb}
\bm{\lambda}^* = \argmin_{\bm{\lambda} \in \mathbb{R}^M} f(\bm{\lambda}) \textrm{ s.t. } \|\bm{\lambda}\|_1 = 1 \textrm{ and } \lambda_i \geq 0 \textrm{ for } i = 1 , \ldots, M,
\end{equation} 
that minimizes the dual cost over all feasible weights, $\bm{\lambda}$.

%% SECTION 4
\section{Solution via subgradient}
\label{sec:solution}
The objective function of \eqref{eq:dualProb} is a nondifferentiable convex function. In this section we show how the subgradient method~\cite{shor2012minimization} can be applied to solve this dual problem. After an appropriate subgradient has been identified, the well-developed literature of subgradient algorithms provides a variety of techniques and step sizes to optimize Problem~\eqref{eq:dualProb} with associated convergence guarantees. 

Recall that a vector $\*g \in \Real^M$ is a subgradient of $f : \Real^M \to \Real$ at $\*x \in \textrm{dom } f$ if for all $\*z \in \textrm{dom } f$,
$$f(\*z)\geq  f(\*x) +  \*g^T(\*z-\*x).$$ In this case we denote that $\*g$ is in the subdifferential of $f$ at $\*x$ by writing $\*g \in \partial f(\*x)$.  If $f$ is differentiable at $\*x$ then the gradient is the only subgradient and $\*g = \nabla f(\*x) = \partial f(\*x).$

To minimize $f$ in Problem~\eqref{eq:dualProb}, the subgradient method uses the iteration
\begin{equation}
\label{eq:subiter}
\bm{\lambda}^{(t+1)} = \Pi(\bm{\lambda}^{(t)} - \alpha^{(t)} \*g^{(t)}),
\end{equation}
where $\alpha^{(t)}$ is a step size selected to guarantee that the sequence $\{ \bm{\lambda}^{(t)}\}_{t=1}^{\infty}$ converges (in distance) to the optimum, $\bm{\lambda}^{*},$ and $\Pi:\Real^M \to \{\*x :  \|\*x\|_1=1, x_i \geq 0 \textrm{ for } i = 1,\ldots,M\} \subset \Real^M$ projects the iterate into the unit simplex.

There is a standard trick for computing a subgradient of the dual function that can adapted to this problem from nonlinear optimization texts such as~\cite{bertsekas1997nonlinear}. Write the Lagrangian as $\mathcal{L}(U,\tau,\bm{\lambda}) = q(U,\tau) + \bm{\lambda}^T\*g(U,\tau),$  where $q(U,\tau)$ is the primal objective function and $\*g(U,\tau) \in \Real^M$ is the vector of constraint values. Given the dual variable, $\bm{\lambda}^{(t)} \in \Real^M,$ at iteration $t,$ let $(U_{\bm{\lambda}^{(t)}},\tau_{\bm{\lambda}^{(t)}})$ be the primal variable that maximizes the Lagrangian. Then $\*g^{(t)} = \*g(U_{\bm{\lambda}^{(t)}},\tau_{\bm{\lambda}^{(t)}})$ is a subgradient of the dual function, $f,$ at $\bm{\lambda}^{(t)}.$

In our case $U_{\bm{\lambda}^{(t)}}$ is defined according by Equation~\eqref{eq:weightedEVD} and the $i$th element of the constraint vector is $g_i(U_{\bm{\lambda}^{(t)}},\tau_{\bm{\lambda}^{(t)}}) = -\tau_{\bm{\lambda}^{(t)}} - \min \{k,p_i\} + \textrm{Tr}(U_{\bm{\lambda}^{(t)}}^TX_i^{}X_i^TU_{\bm{\lambda}^{(t)}}^{}).$ However, the constant vector $[-\tau_{\bm{\lambda}^{(t)}}, \ldots, -\tau_{\bm{\lambda}^{(t)}}]^T \in\Real^M$ does not affect the direction after projection onto the unit simplex, so a subgradient of $f(\bm{\lambda}^{(t)})$ is
\begin{equation}
\label{eq:generic_subgrad}
\*g^{(t)} =
\begin{pmatrix}
- \min \{k,p_1\} + \textrm{Tr}(U_{\bm{\lambda}^{(t)}}^TX_1^{}X_1^TU_{\bm{\lambda}^{(t)}}^{}) \\
\vdots \\
- \min \{k,p_M\} + \textrm{Tr}(U_{\bm{\lambda}^{(t)}}^TX_M^{}X_M^TU_{\bm{\lambda}^{(t)}}^{})
\end{pmatrix}.
\end{equation}
We can check that $\*g^{(t)}$ is a subgradient of $f$ as follows. For any $\tilde{\bm{\lambda}} \in \Real^M$ such that $\| \tilde{\bm{\lambda}}\|_1 = 1$ and $\tilde{\lambda}_i \geq 0 $ for $i = 1, \ldots, M$  we have
\begin{equation}
\begin{aligned}
f(\bm{\lambda}^{(t)}) + \*g^{(t)T}(\tilde{\bm{\lambda}} - \bm{\lambda}^{(t)}) &= f(\bm{\lambda}^{(t)}) + \*g^{(t)T}\tilde{\bm{\lambda}}-\*g^{(t)T}\bm{\lambda}^{(t)} \\
&= f(\bm{\lambda}^{(t)}) + \*g^{(t)T}\tilde{\bm{\lambda}}  - f(\bm{\lambda}^{(t)}) \\
&= -\sum_{i=1}^M \tilde{\lambda_i^{}} \min \{k,p_i\}  + \textrm{Tr}(U_{\bm{\lambda}^{(t)}}^T(\sum_{i=1}^M \tilde{\lambda_i^{}} X_i^{}X_i^T ) U_{\bm{\lambda}^{(t)}}^{})\\
&\leq -\sum_{i=1}^M \tilde{\lambda_i^{}} \min \{k,p_i\}  + \max_{U^TU = I}  \textrm{Tr}(U^T(\sum_{i=1}^M \tilde{\lambda_i^{}}X_i^{}X_i^T ) U)\\
&= f(\tilde{\bm{\lambda}}),
\end{aligned}
\end{equation}
and thus $\*g^{(t)} \in \partial f(\bm{\lambda}^{(t)}).$ 

\subsection{Convergence}
\label{subsec:conv}
The subgradient $\*g^{(t)}$ can be used to update $\bm{\lambda}^{(t)}$ via the iteration in~\eqref{eq:subiter}. The subgradient method is not a descent method, so the value of the objective function at step $t+1$ may be larger than it was at step $t$. Thus we keep track of the dual variable with the lowest cost at each iteration and denote it
\begin{equation}
\bm{\lambda}^{(t+1)}_{\textrm{best}} = 
\begin{dcases} 
      \bm{\lambda}^{(t)}_{\textrm{best}} & f(\bm{\lambda}^{(t+1)}) > f(\bm{\lambda}_{\textrm{best}}^{(t)}); \\[4pt]
      \bm{\lambda}^{(t+1)} & \textrm{otherwise.}
\end{dcases}
\end{equation}

Given an upper bound on the norm of the subgradients, $\|g^{(t)}\|_2 \leq G < \infty$ for all $t,$ classical theory makes different guarantees on the convergence of the sequence of iterates, $\{\bm{\lambda}^{(t)}\}_{t=1}^{\infty},$  and thus on the sequence of objective function values, $\{f(\bm{\lambda}^{(t)}_{\textrm{best}} )\}_{t=1}^{\infty},$ depending on the choice of step size, $\alpha^{(t)}.$  For example, with step sizes independent of iteration like $\alpha^{(t)} = a$ or $\alpha^{(t)} = \nicefrac{a}{\|\*g^{(t)}\|_2}$ for some $a>0$, the subgradient algorithm will converge respectively to within $\nicefrac{G^2 a }{2}$ or $\nicefrac{G a }{2}$ of the optimal value~\cite{bertsekas1997nonlinear}. Alternatively, if the step size converges to zero and the sequence is nonsummable or square-summable, that is, $\lim_{t \to \infty} \alpha^{(t)} = 0$ and
\begin{equation}
\label{eq:decreasingStep}
\sum_{t=1}^{\infty} \alpha^{(t)} = \infty \quad \textrm{or} \quad \sum_{t=1}^{\infty} (\alpha^{(t)})^2 < \infty,
\end{equation}
the subgradient method converges to an optimal objective value, $\lim_{t \to \infty} f(\bm{\lambda}^{(t)}_{\textrm{best}}) = f(\bm{\lambda}^*).$ These conditions are satisfied by step sizes like, $\alpha^{(t)} = \nicefrac{a}{\sqrt{t}}$ for $a>0,$ or $\alpha^{(t)} = \nicefrac{a}{(b+t)}$ where $a > 0$ and $b \geq 0.$ Proofs of these results can be found in standard literature on convex optimization for nonsmooth problems such as \cite{bertsekas1997nonlinear,shor2012minimization, hiriart2013convex}.

Although the theory requires $\alpha^{(t)}$ to satisfy the constraints in \eqref{eq:decreasingStep} for convergence, the small step size leads to very slow convergence.  In practice we can find an approximate solution quickly by stepping in the direction of a subgradient but requiring the dual objective to decrease at each iteration. Algorithm~\ref{alg:subgrad} (in Appendix~\ref{sec:alg}) solves Problem~\eqref{eq:dualProb} by performing a back-tracking line search in the direction of $\*g^{(t)} \in \partial f(\bm{\lambda}^{(t)})$ to ensure that the dual objective decreases at each step, however, this method is not guaranteed to converge because $\*g^{(t)}$ is not necessarily a descent direction.  The practical implementation of Algorithm~\ref{alg:subgrad} is a hybrid of a back-tracking line search and a nonsummable diminishing step size and for a fixed dimension $k$ it identifies a stationary point of the dual problem while providing a feasible solution to the primal problem. It is not intended to be a state-of-the-art subgradient algorithm, but rather just one example of an implementation that is faster than the standard $\nicefrac{a}{(b+t)}$ square-summable step size. Alternatively, a well-established quasi-Newton method like the Broyden-Fletcher-Goldfarb-Shanno (BFGS) algorithm~\cite{curtis2017bfgs} can be used to solve Equation~\eqref{eq:dualProb}, but empirically the convergence rates are comparable to those of the algorithm presented here for this problem.

\subsection{Optimality}
\label{subsec:opt} 
In addition to theoretical convergence guarantees, the optimality of a solution to the dual subgradient approach can be verified in some cases. Let $\bm{\lambda}^{*}$ be a solution to Problem~\eqref{eq:dualProb}. There exists a matrix $U_{\bm{\lambda}^{*}}$ whose columns are the $k$ dominant eigenvectors of $\sum_{i=1}^M \lambda_i^{*}X_i^{} X_i^{T}$, analogous to Equation~\eqref{eq:weightedEVD}. Then $U_{\bm{\lambda}^{*}}$ satisfies $U_{\bm{\lambda}^{*}}^T U_{\bm{\lambda}^{*}} = I$ and is thus a feasible solution to the primal problem in \eqref{eq:matrix_eq}. If the primal and dual objective functions are equal, strong duality holds and implies that $\bm{\lambda}^{*}$ and $\*U^{*} = \textrm{col}(U_{\bm{\lambda}^{*}})$ are globally optimal dual and primal variables, respectively. Empirically this occurs for collections of data that satisfy an implicit assumption of minimax optimization; that the data collection is free of outliers.  Even when strong duality does not hold, the duality gap gives a bound on the maximum possible improvement for a solution. 

This verification of optimality is standard for problems where the primal and dual costs are both computable, but existing techniques for finding the GMEB do not offer this feature. For instance, using a primal method like \cite{renard2018grassmannian} does not directly provide a solution to the dual problem, and thus the duality gap is unknown. Section~\ref{subsec:accuracy} contains numerical experiments that demonstrate the accuracy of the proposed subgradient method. 

%% SECTION 5
\section{Proposed order selection rule}
\label{sec:ord_select}
Given a dimension, $k$, and a finite collection of subspaces, $\mathcal{D} = \left\{\*X_i \in \textrm{Gr}(p_i,n) \right\}_{i=1}^M,$  there exist subspaces,
\begin{equation}
\*{U}^{*}(k)  = \underset{\*U \in \textrm{Gr}(k,n)}{\argmin} \ \underset{i = 1, \ldots, M}{\max} d_{\textrm{Gr}(k,n)}(\*U,\*X_i),
\label{eq:minmax_cost}
\end{equation}
for $k = 1, \ldots \max_i\{\textrm{dim}(\*X_i)\}$. The argument $k$ is now included in the notation for the GMEB center to emphasize that the subspace depends on the parameter $k$, and may differ significantly depending on the value of this parameter. Section~\ref{sec:solution} described a method to compute $\*U^*(k)$ from the associated dual variable, $\bm{\lambda}^{*}(k) \in \Real^M.$   However, because $\mathcal{D}$ contains subspaces of differing dimension, it is unclear on which Grassmannian the minimum enclosing ball should be computed. Thus, given the set $\mathcal{D}$, in this section we would like to determine the optimal choice for $k,$ in addition to the associated center $\*U^*(k)$. Please note a change in notation; the costs associated with a particular order, $k$, are more intuitive when the primal is formulated as a minimization problem and the dual is a maximization. Therefore, as shown in Equation~\eqref{eq:minmax_cost}, the primal minimization formulation is used for the remainder of the manuscript. The prior formulation was only used for ease of notation in the subgradient method. 

All orthogonally invariant distances on Gr$(k,n)$ can be written as a function of the $k$ principle angles between a pair of points. It should be clear from the definition in Equation~\eqref{eq:angles} that each angle is bounded above by $\nicefrac{\pi}{2},$ and thus that the squared chordal distance is bounded above by $k$. Scaling the primal objective function by $\nicefrac{1}{k}$ normalizes the cost associated with $\*U^{*}(k)$ so that the value of
\begin{equation}
\label{eq:scaled_obj}
c_{\textrm{obj}}(k):=  
\begin{dcases}
0 & k=0; \\
\max_{i = 1, \ldots, M} \frac{d_{\textrm{Gr}(k,n)}(\*U^{*}(k),\*X_i)}{k} & k = 1 , \ldots \max_i\{\textrm{dim}(\*X_i)\},
\end{dcases}
\end{equation}
gives a fair comparison across different values of $k.$ The normalized objective function achieves its maximum value, $c_{\textrm{obj}}(k)=1,$ when there exits an $i$ such that $\*X_i \perp \*U^{*}(k).$ That is, $\*U^{*}(k)$ contains no information about at least one of the points in $\mathcal{D}$. At the other extreme, the minimum occurs when $k=0$, and when the point of each $\Omega_{*}(\*X_i)$ closest to the center coincides with the center.  That is, $c_{\textrm{obj}}(k)=0$ when $\*Y^{*}_i(k) = \*U^{*}(k)$ for all $i,$ where $\*Y_i^{*}(k) = \underset{\*Y_i \in \Omega_{*}(\*X_i)}{\argmin} d_{\textrm{Gr}(k,n)}(\*U^{*}(k),\*Y_i).$ 

Simply minimizing $c_{\textrm{obj}}(k)$ with respect to $k$ is not sufficient to identify the ideal dimension of $\*U^{*}(k)$ because on average $c_{\textrm{obj}}(k) \leq c_{\textrm{obj}}(k+1)$ irrespective of the relationship between the data points, and of course $c_{\textrm{obj}}(0) = 0$ by definition.  However, the dimension of the ideal center should represent all the common information without over-fitting, and should also indicate when no significant relationship exists between the data. Thus we propose a penalty term based on the dimensions of the data not represented by $\*U^{*}(k)$ that balances the information lost by making $k$ too small with the lack of specificity that comes from setting $k$ too large.  

Let $\*U^{*\perp}(k)$ denote the orthogonal complement of $\*U^{*}(k)$ and $\tilde{p}_j \doteq \min\{n-k,\textrm{dim}(\*X_j)\}$ for $j=1,\ldots,M.$ The expression
\begin{equation}
\label{eq:proposed_penalty}
c_{\textrm{pen}}(k):=  
\begin{dcases}
1 & k=0; \\
\underset{j=1,\ldots,M}{\min} 1 -\frac{ d_{\textrm{Gr}(\tilde{p}_j,n)}(\*U^{*\perp}(k),\*X_j)}{\tilde{p}_j} & k = 1 , \ldots \max_j\{\textrm{dim}(\*X_j)\},
\end{dcases}
\end{equation}
represents the minimum similarity between any point in $\mathcal{D}$ and the dimensions not contained in the center of the GMEB. A high minimum similarity between points in $\mathcal{D}$ and $\*U^{*\perp}(k)$ implies that too much information is being left out of the central subspace, $\*U^{*}(k)$.  The penalty term takes a value of $c_{\textrm{pen}}(k) = 1$ when dim$(\*U^{*\perp}(k) \cap \*X_j) = \tilde{p}_j$ for all $j$ and $c_{\textrm{pen}}(k) = 0$ when there exists a $j$ for which $\*X_j \perp \*U^{*\perp}(k).$ The sum of the terms in \eqref{eq:scaled_obj} and \eqref{eq:proposed_penalty} leads to the proposed order selection rule, 
\begin{equation}
\label{eq:order_rule}
k^{*} = \argmin_{k= 0,\ldots, \max_i\{\textrm{dim}(\*X_i)\}}  c_{\textrm{obj}}(k) + c_{\textrm{pen}}(k).
\end{equation}
The two terms in \eqref{eq:order_rule} are computed independently so the GMEB center is not affected by the penalty term. The value of $k^{*}$ that minimizes the sum of these two terms corresponds to the number of subspace dimensions needed to represent the common information present in $\mathcal{D}$ without over-fitting.  Numerical experiments in Section~\ref{subsec:order_selection} demonstrate the efficacy of the order selection rule on simulated data with ground truth.

\subsection{Primal solutions are not nested in general for increasing values of $k$}
\label{subsec:not_nested}
Naively, the order selection rule in Equation~\eqref{eq:order_rule} can be applied by computing the costs $c_{\textrm{obj}}(k)$ and $c_{\textrm{pen}}(k)$ independently for $k=0,\ldots, \max_i\{\textrm{dim}(\*X_i)\}$ as follows,
\begin{enumerate}
\item Compute $\bm{\lambda}^{*}(k)$ using the subgradient method described in Section~\ref{sec:solution}.
\item Find the associated primal variable, $\*U^{*}(k),$ as the $k$-dimensional eigenspace of the weighted sum $\sum_{i=1}^M \lambda_i^{*}(k) X_i^{} X_i^T.$
\item Compute the orthogonal complement, $\*U^{*\perp}(k) =  \textrm{col}\left(I - U^{*}(k) U^{*T}(k)\right).$ 
\end{enumerate}
Then $k^{*}$ is selected as the value of $k$ associated with the minimum cost, $c_{\textrm{obj}}(k) + c_{\textrm{pen}}(k)$. If $\bm{\lambda}^{*}(k) = \bm{\lambda}^{*}(k+1)$ for some $k<\max_i\{\textrm{dim}(\*X_i)\}$ then the solution on Gr$(k+1,n)$ can be constructed in a greedy fashion as the direct sum of the solution on Gr$(k,n)$ and the $(k+1)$st eigenvector of $\sum_{i=1}^M \lambda_i^{*}(k) X_i^{} X_i^T.$ Unfortunately, the dual variables are not generally equal for increasing values of $k$, so a greedy approach is not appropriate. 

Observe that the central subspaces are not nested for increasing dimensions in the following illustrative example. Let 
\begin{equation}
\begin{aligned}
\label{eq:example_pts}
X_1 = 
\begin{bmatrix} 
\frac{\sqrt{2}}{\sqrt{3}}& 0 \\
\frac{1}{\sqrt{6}}& 0 \\
\frac{1}{\sqrt{6}} & 0 \\
0 & \frac{\sqrt{7}}{\sqrt{8}} \\
0 & \frac{1}{\sqrt{8}}
\end{bmatrix},  & &
X_2 = 
\begin{bmatrix} 
\frac{1}{\sqrt{6}}& 0 \\
\frac{\sqrt{2}}{\sqrt{3}}& 0 \\
\frac{1}{\sqrt{6}} & 0 \\
0 & \frac{1}{\sqrt{8}} \\
0 & \frac{\sqrt{7}}{\sqrt{8}}
\end{bmatrix},
 & & \textrm{ and }
X_3 = 
\begin{bmatrix} 
\frac{1}{\sqrt{6}} \\
\frac{1}{\sqrt{6}} \\
\frac{\sqrt{2}}{\sqrt{3}}   \\
0  \\
0 
\end{bmatrix},
\end{aligned}
\end{equation}
be orthonormal bases for the three points $\*X_1, \*X_2 \in \textrm{Gr}(2,5) \textrm{ and } \*X_3 \in \textrm{Gr}(1,5).$ One can check that the subspace that minimizes the maximum distance to these three points on Gr$(1,5)$ is the mean of their first columns. That is, the optimal primal and dual variables are
\begin{equation}
\begin{aligned}
\*U^{*}(1) = \textrm{col}\left(
\begin{bmatrix} 
\frac{1}{\sqrt{3}}&
\frac{1}{\sqrt{3}}&
\frac{1}{\sqrt{3}} &
0  &
0
\end{bmatrix}^T\right),
& & \textrm{ and }
\bm{\lambda}^{*}(1) = 
\begin{bmatrix} 
\frac{1}{\sqrt{3}}&
\frac{1}{\sqrt{3}}&
\frac{1}{\sqrt{3}}
\end{bmatrix}^T,
\end{aligned}
\end{equation}
with associated primal and dual costs of 
\begin{equation}
\label{eq:kOne_duality_gap}
\underset{\*U \in \textrm{Gr}(1,5)}{\min} \max_{i = 1, 2, 3} d_{\textrm{Gr}(1,5)}(\*U,\*X_i) = \max_{\bm{\lambda} \in \mathbb{R}^3} \min_{U^TU = I}  1 - \sum_{i=1}^3 \lambda_i^{} \textrm{Tr}(U^{T}Y_i^{} Y_i^{T}U^{}) = \frac{1}{9}.
\end{equation} 
The duality gap in Equation~\eqref{eq:kOne_duality_gap} is zero, indicating that this is a global solution.

On Gr$(2,5)$, however, $\Omega_{+}(\*X_3)$ consists of subspaces that span $X_3$ and any orthogonal direction. In particular there exists $\*Y_3 \in \Omega_{+}(\*X_3)$ such that the second column of $Y_3$ is $\left[0 \ 0 \ 0 \ \nicefrac{1}{\sqrt{2}} \ \nicefrac{1}{\sqrt{2}}\right]^T.$ This leads to a solution for the center of the minimum enclosing ball on Gr$(2,5)$ given by primal and dual variables
\begin{equation}
\begin{aligned}
\*U^{*}(2) = \textrm{col} \left(
\begin{bmatrix} 
\frac{3}{\sqrt{22}}& \frac{3}{\sqrt{22}}&  \frac{2}{\sqrt{22}} & 0 & 0 \\
0 & 0 & 0 & \frac{1}{\sqrt{2}}  & \frac{1}{\sqrt{2}}
\end{bmatrix}^T\right), && \textrm{ and }
\bm{\lambda}^{*}(2) = 
\begin{bmatrix} 
\frac{1}{2} & \frac{1}{2} & 0 
\end{bmatrix}^T.
\end{aligned}
\end{equation}
Notably, $\*X_3$ is not in the support of the minimum enclosing ball on Gr$(2,5)$ and thus does not influence the central subspace. Strong duality also holds for this solution with 
\begin{equation}
\label{eq:kTwo_duality_gap}
\underset{\*U \in \textrm{Gr}(2,5)}{\min} \max_{i = 1, 2, 3} d_{\textrm{Gr}(2,5)}(\*U,\*X_i) = \max_{\bm{\lambda} \in \mathbb{R}^3} \min_{U^TU = I}  2 - \sum_{i=1}^3 \lambda_i^{} \textrm{Tr}(U^{T}Y_i^{} Y_i^{T}U^{}) = \frac{14-3\sqrt{7}}{24}.
\end{equation} 
Since $\*U^{*}(1)$ is orthogonal to the second dimension of $\*U^{*}(2)$ and noncollinear with the first, and the columns of $U^{*}(2)$ are orthogonal, we have $\*U^{*}(1) \not\subset \*U^{*}(2).$ Additionally we find that optimal order selected by applying the rule in Equation~\eqref{eq:order_rule} is $k^* = 1,$  because 
\begin{equation}
\begin{aligned}
&c_{\textrm{obj}}(0) + c_{\textrm{pen}}(0) = 0 + 1 = 1, \\
&c_{\textrm{obj}}(1) + c_{\textrm{pen}}(1) = \frac{1}{1}\left(\frac{1}{9}\right)+ \frac{1}{1}\left( 1 - \left(\frac{\sqrt{8}}{\sqrt{9}}\right)^2\right)\approx 0.22, \ \textrm{ and}\\
&c_{\textrm{obj}}(2) + c_{\textrm{pen}}(2) = \frac{1}{2}\left(\frac{14-3\sqrt{7}}{24}\right) + \frac{1}{2}\left(2 - \left( \frac{-1}{\sqrt{12}}\right)^2 + \left(\frac{1 - \sqrt{7}}{\sqrt{16}} \right)^2\right)\approx 0.25.
\end{aligned}
\end{equation}
This agrees with the intuition that the center of the minimum enclosing ball represents the common information in all points without over-fitting to any subset of points, but note that the optimal order is not always the dimension of the smallest subspace. The common subspace may have dimension smaller than any of the samples or there may be no common subspace.

Even though the primal solutions are not always nested, a good initial guess for the dual variable will reduce computational overhead. One benefit of the subgradient approach is that $\bm{\lambda}^{*}(k)$ is computed explicitly.  Thus we can initialize the algorithm with $\bm{\lambda}^{(0)}(k+1) = \bm{\lambda}^{*}(k)$. The impact of this heuristic warm-start is discussed in the experiments in Section~\ref{subsec:warm_start}.

\subsection{Related literature on order fitting for subspace averaging}
A recent work from Santamar{\'\i}a \etal~\cite{santamaria2016order} also attempts to find a central subspace of ambiguous dimension. The authors minimize the mean-squared error (MSE) between a subspace and a collection of data in the space of $n \times n$ projection matrices using the squared Frobenius norm. That is,
\begin{equation}
\label{eq:mse}
E(k) = \underset{\*U \in \textrm{Gr}(k,n)}{\min}\frac{1}{M}\sum_{i=1}^M \|U U^{T} - X_i^{} X_i^T \|_F^2.
\end{equation}

Putting aside for a moment that the current work is interested in minimizing the maximum deviation rather than the mean-squared error, there remains a central difference between the technique in~\cite{santamaria2016order} and the proposed method. The optimization of Equation~\eqref{eq:mse} is done in a vector space, after which the solution is mapped to the nearest point on the Grassmann manifold. This is subtly different than minimizing the MSE on the Grassmannian with respect to the squared chordal distance using the point-to-set interpretation of~\cite{ye_lim}.  To see this, write half of the squared distance from \cite{santamaria2016order} between the central subspace and the $i$th point as
\begin{equation}
\label{eq:santa_dist}
\begin{aligned}
 \frac{1}{2}\|U^{*}(k) U^{*T}(k)  - X_i^{} X_i^T \|_F^2 &= \frac{k + p_i}{2} - \sum_{r=1}^{\min\{k,p_i\}} \cos^2 (\theta_r(\*U^{*}(k) ,\*X_i))\\
&= \frac{|k - p_i|}{2} + \sum_{r=1}^{\min\{k,p_i\}} \sin^2 (\theta_r(\*U^{*}(k) ,\*X_i)).
\end{aligned}
\end{equation} 
In contrast, the point-to-set squared chordal distance on $\textrm{Gr}(k,n)$ is
\begin{equation}
\label{eq:chordal}
\begin{aligned}
d_{\textrm{Gr}(k,n)}(\*U^{*}(k) ,\*X_i) &= \min \big\{ d(\*U^{*}(k) ,\*Y_i) : \*Y_i \in \Omega_{*}(\*X_i)\big\} \\
&= \min \big\{\frac{1}{2}\|U^{*}(k)  U^{*T}(k)  - Y_i^{} Y_i^T \|_F^2 : \*Y_i \in \Omega_{*}(\*X_i)\big\}  \\
&= k - \sum_{r=1}^{k} \cos^2 (\theta_r(\*U^{*}(k) ,\*Y_i))\\
%&= \sum_{r=1}^k \sin^2 (\theta_r(\*U^{*}(k) ,\*Y_i)) \\
&=\sum_{r=1}^{\min \{k,p_i\}} \sin^2 (\theta_r(\*U^{*}(k) ,\*X_i)) \\
\end{aligned} 
\end{equation} 
because $0 = \theta_{p_i}(\*U^{*}(k) ,\*Y_i) = \theta_{p_i+1}(\*U^{*}(k) ,\*Y_i) = \cdots = \theta_{k}(\*U^{*}(k) ,\*Y_i)$ if $p_i<k$ by the definition of $\*Y_i$ in Equation~\eqref{eq:y_def}.  Thus the distances differ by $\frac{|k-p_i|}{2},$ which is the difference in dimensions between the central subspace and the $i$th data point. 

The slight difference in distance measurements lends itself to an interesting interpretation when determining the appropriate rank of the central subspace. The solution to 
\begin{equation}
\label{eq:flag}
\*U^{*}(k) = \underset{\*U \in \textrm{Gr}(k,n)}{\argmin}\frac{1}{M}\sum_{i=1}^M \|U^{} U^{T} - X_i^{} X_i^T \|_F^2 %d^2(\*U,\*X_i).
\end{equation}
for a fixed $k$ is the dominant $k$-dimensional eigenspace of the sum $\frac{1}{M}\sum_{i=1}^M X_i^{} X_i^T$.  That is, if
\begin{equation}
\label{eq:decomp}
\frac{1}{M}\sum_{i=1}^M X_i^{} X_i^T = F D F^T
\end{equation} is an eigendecomposition with eigenvectors $F = [\*f_1, \*f_2, \ldots, \*f_{R}]$ and associated eigenvalues $d_{1} \geq d_{2} \geq \cdots \geq d_{R},$ then the solution to Equation~\eqref{eq:flag} is $\*U^{*}(k) = [\*f_1, \*f_2, \ldots, \*f_k].$ Note that this $\*U^{*}(k)$ is not the same subspace as the center of the minimum enclosing ball. The MSE in Equation~\eqref{eq:mse} can be written as a function of all $R$ eigenvalues,
\begin{equation}
\label{eq:mse_eigs}
E(k) = \sum_{r=1}^k 1-d_r + \sum_{r=k+1}^R d_r,
\end{equation}
and the minimum of Equation~\eqref{eq:mse_eigs} is achieved when $k^{*}$ is the smallest value for which $d_{k+1} < 0.5$. This eigenvalue threshold is then fixed regardless of the dimension of the ambient space, and as we will see in Section~\ref{subsec:order_selection}, the selected dimension could differ drastically for noisy data depending on the ambient dimension.

For a different interpretation of the $k^{*}$ that minimizes Equation~\eqref{eq:mse} we can rewrite Equation~\eqref{eq:mse_eigs} as a function of the angles between each eigenvector and the subspaces,
\begin{eqnarray}
E(k) &=& \sum_{r=1}^k 1-\*f_r^T(\frac{1}{M}\sum_{i=1}^M X_i^{} X_i^T)\*f_r + \sum_{r=k+1}^R \*f_r^T(\frac{1}{M}\sum_{i=1}^M X_i^{} X_i^T)\*f_r\\
&=& \sum_{r=1}^k 1 - \frac{1}{M}\sum_{i=1}^M \cos^2(\theta(\*f_r,\*X_i)) + \sum_{r=k+1}^R \frac{1}{M}\sum_{i=1}^M \cos^2(\theta(\*f_r,\*X_i))\\
&=&\sum_{r=1}^k \frac{1}{M}\sum_{i=1}^M \sin^2(\theta(\*f_r,\*X_i)) + \sum_{r=k+1}^R \frac{1}{M}\sum_{i=1}^M \sin^2(\frac{\pi}{2}- \theta(\*f_r,\*X_i)) \label{eq:avg_angles3}\\
&=&\sum_{r=1}^k \frac{1}{M}\sum_{i=1}^M \sin^2(\theta(\*f_r,\*X_i)) + \sum_{r=k+1}^R \frac{1}{M}\sum_{i=1}^M \sin^2(\theta(\*f_r,\*X_i^{\perp})) \label{eq:avg_angles4} \\
&=&\sum_{r=1}^k \frac{1}{M}\sum_{i=1}^M d_{\textrm{Gr}(1,n)}(\*f_r,\*X_i) + \sum_{r=k+1}^R \frac{1}{M}\sum_{i=1}^M d_{\textrm{Gr}(1,n)}(\*f_r,\*X_i^{\perp}). \label{eq:avg_angles5}
\end{eqnarray}
The equality between \eqref{eq:avg_angles3} and \eqref{eq:avg_angles4} is due to~\cite[Thm.~2.7]{knyazev2006majorization} which implies that $\frac{\pi}{2}- \theta(\*f_r,\*X_i) = \theta(\*f_r,\*X_i^{\perp}).$ Note, however, that Equation~\eqref{eq:avg_angles5} is \textit{not} equivalent to 
\begin{equation}
\frac{1}{M}\sum_{i=1}^M d_{\textrm{Gr}(k,n)}(\*U^{*}(k),\*X_i) +  \frac{1}{M}\sum_{i=1}^M d_{\textrm{Gr}(R-k,n)}(\*U^{*\perp}(k),\*X_i^{\perp})
\end{equation}
because linear combinations of the eigenvectors, $\*f_r,$ are not included in the expression. A new interpretation of the MSE-minimizing $k$ becomes fairly apparent in light of Equation~\eqref{eq:avg_angles5}. The optimal $k^{*}$ is the one that minimizes the mean-squared chordal distance between $\left\{ \*f_{1}, \ldots, \*f_k\right\}$ and the data points, plus the mean-squared chordal distance between $\left\{ \*f_{k+1}, \ldots, \*f_R \right\}$ and the orthogonal complements of the data points.

\subsection{Hybrid rule}
\label{subsec:hybrid}
It is possible to create a hybrid of the order-selection rule of~\cite{santamaria2016order} and the proposed method with a slight modification. 
In~\cite{garg2019subspace}, a robustification of the technique in~\cite{santamaria2016order} is proposed that leads to a weighted eigenvalue decomposition at optimality.  The weights are determined using a variety of robust objective functions via a majorization-minimization scheme, which results in a down-weighting of outliers in the data.  By minimizing the mean-squared error of the \textit{weighted} average (similar to Equation~\eqref{eq:mse}), this amounts to a hard eigenvalue threshold with the order chosen to be the number of dimensions with eigenvalues greater than $0.5$.  

For the hybrid method, weights will come from the values of the dual variable, $\bm{\lambda}^{*}(k),$ at optimality. Since these values depend on the parameter $k,$ the hard eigenvalue threshold is not applicable. Let $d_1(k) \geq d_2(k) \geq \cdots \geq d_R(k)$ be the eigenvalues of $\sum_{i=1}^{M} \lambda^{*}_i(k) X_i^{} X_i^T$ where $\bm{\lambda}^{*}(k)$ is the vector of optimal dual variables computed for the GMEB on Gr$(k,n)$ using the proposed algorithm. For $k=0,$ let $\lambda_i^{*}(0) =\frac{1}{M}$ for $i=1,\ldots,M.$ We define a modified version of the MSE from Equation~\eqref{eq:mse_eigs} as
\begin{equation}
\label{eq:modified_mse}
\tilde{E}(k) = \sum_{r=1}^k 1-d_r(k) + \sum_{r=k+1}^R d_r(k).
\end{equation}
The order-selection rule of~\cite{santamaria2016order} applied to the GMEB center is then
\begin{equation}
\label{eq:modified_santa_rule}
k^* = \underset{k=0,\ldots, \max_i\{\textrm{dim}(\*X_i)\}}{\argmin}\tilde{E}(k).
\end{equation}
It should be clear that the eigenvalues $\{d_r(k)\}_{r=1}^R$ will be different for different values of $\bm{\lambda}^{*}(k).$ In the experiments of Section~\ref{subsec:order_selection}, this combined method is referred to as ``Hybrid'' and performs favorably for all tests; out-performing the other techniques in $2$ out of $3$ scenarios.

%% SECTION 6
\section{Synthetic data generation}
\label{sec:data}
The numerical experiments in Section~\ref{sec:numerical} require data for which the ground truth is known, and ideally data for which the center of the GMEB is distinct from the other generalized Grassmannian means. Thus, in this section we propose two different models for sampling points nonuniformly from a unit ball on the Grassmannian.  The first is an asymmetrical nested ball structure, and the second samples more densely within a randomly selected arc of the boundary of a unit ball. 

\subsection{Asymmetrical nested ball model}
\label{subsec:nested}
A collection of subspaces, $\mathcal{D} = \{\*X_i\}_{i=1}^{M},$ are uniformly sampled from two balls, $\mathcal{B}_{\epsilon_2}(\*Z_2) \subset \mathcal{B}_{\epsilon_1}(\*Z_1) \subset \textrm{Gr}(k_0,n)$ with centers at $\*Z_1, \ \*Z_2$ and corresponding radii $\epsilon_1 > \epsilon_2,$ respectively. The larger ball, $\mathcal{B}_{\epsilon_1}(\*Z_1),$ is the minimum enclosing ball of the data so that $\*U^*(k_0) = \*Z_1$.  The smaller ball is fully contained within the larger ball, i.e., $\mathcal{B}_{\epsilon_2}(\*Z_2) \subset \mathcal{B}_{\epsilon_1}(\*Z_1),$ but $\*Z_1 \notin \mathcal{B}_{\epsilon_2}(\*Z_2)$. Let $M_1,M_2$ be the number of points sampled from $\mathcal{B}_{\epsilon_1}(\*Z_1),\mathcal{B}_{\epsilon_2}(\*Z_2)$ respectively, with $M = M_1 + M_2$. When $M_2 = 0$, the generalized Grassmannian means are all equal to the point $\*Z_1$. When more points are sampled from $\mathcal{B}_{\epsilon_2}(\*Z_2)$ and the fraction $\nicefrac{M_2}{M_1}$ grows, the generalized Grassmannian means for $p < \infty$ move away from $\*Z_1$ in the direction of $\*Z_2$, making the averages distinct without affecting the center of the GMEB. The radius of the large ball, $\epsilon_1,$ controls the similarity of the data points.
\begin{figure*}[!t]
\centering
\input{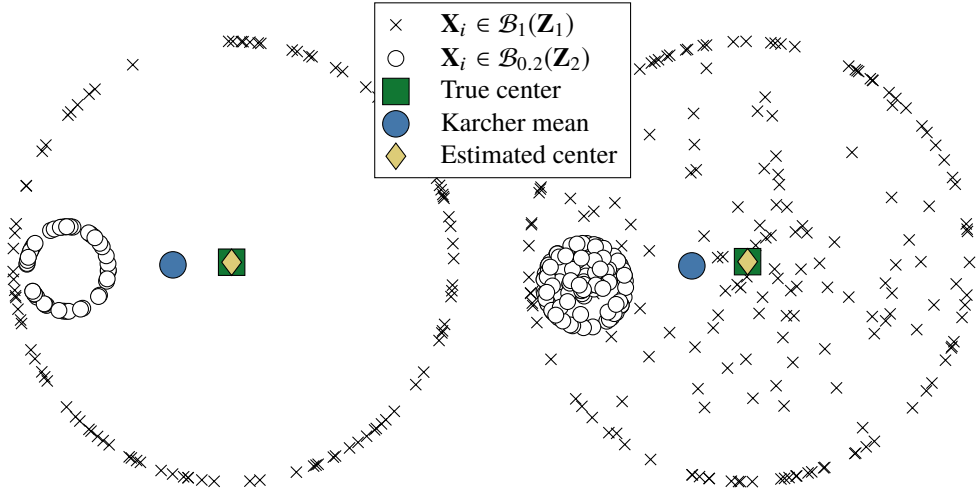}
\caption{Two examples of point sets from Gr$(1,3)$ generated using the nested ball model embedded into $\Real^2$ by multidimensional scaling. The points from $\mathcal{B}_1(\*Z_1)$ are indicated with x's, points from $\mathcal{B}_{0.2}(\*Z_2)$ are marked with white circles, the true center is the green square, the Karcher mean is the blue circle, and the estimated GMEB center is the yellow diamond. }
\label{fig:3d}
\end{figure*}

As described, the data points are all sampled from a single manifold, Gr$(k_0,n).$ If $\epsilon_1$ is small enough, then the optimal rank for the GMEB (or any of the generalized Grassmannian means) is $k^* = k_0$. This construction can be generalized in two ways.
\begin{enumerate}
\item For $i=1,\ldots,M$, the basis for $\*X_i$ can be completed to a $p_i$-dimensional subspace by taking the span of $X_i$ and $p_i-k_0$ random dimensions. If the $p_i-k_0$ random dimensions are mutually orthogonal for $i=1,\ldots, M$, then the optimal rank for the GMEB is still $k^* = k_0$.
\item Points from the large ball can be sampled from one manifold, $\mathcal{B}_{\epsilon_1}(\*Z_1) \subset \textrm{Gr}(k_1,n)$ while points from the small ball are sampled from another, $\mathcal{B}_{\epsilon_2}(\*Z_2) \subset \textrm{Gr}(k_2,n).$ If $k_1 \neq k_2$, the optimal rank of the central subspace is ambiguous. Experiments show that using the proposed order selection rule, $k^* = k_1$ independent of other parameters, but using the criteria of \cite{santamaria2016order}, $k^*$ depends on $\epsilon_1$ and $\nicefrac{M_2}{M_1}$.
\end{enumerate}
As an illustrative example, Figure~\ref{fig:3d} shows $2$-dimensional embeddings via multidimensional scaling of data sets on Gr$(1,3)$ that have been generated according to the asymmetrical nested ball model. The yellow diamond indicates the center of the GMEB (computed via the proposed method) and the blue circle marks the Karcher mean of each data collection.

\subsection{Unit ball with higher sampling density from a random arc}
\label{subsec:nonuni}
Another practical scenario where the GMEB center may differ from other generalized Grassmannian means is when data has been sampled unevenly. This setting is simulated by selecting a random arc from the boundary of a unit ball and sampling additional points from that region. A collection of subspaces, $\mathcal{D} = \{\*X_i\}_{i=1}^{M},$ are uniformly sampled from the ball $\mathcal{B}_{\epsilon_1}(\*Z_1) \subset \textrm{Gr}(k_0,n)$ with center at $\*Z_1$ and radius $\epsilon_1$.  $M_1$ points are sampled from $\mathcal{B}_{\epsilon_1}(\*Z_1)$ so that $\*U^*(k_0) = \*Z_1$. Two points are randomly selected from the boundary of $\mathcal{B}_{\epsilon_1}(\*Z_1),$ and $M_2$ additional points are uniformly sampled from the arc connecting them on the boundary to create $M = M_1 + M_2$ samples. The data points are all sampled from a single manifold, Gr$(k_0,n),$ and for sufficiently small $\epsilon_1,$ the optimal rank for the GMEB (or any of the generalized Grassmannian means) is $k^* = k_0$. To generalize this construction, additional dimensions can be included to create points from a disjoint union of Grassmannians.
\begin{figure*}[!t]
\centering
\input{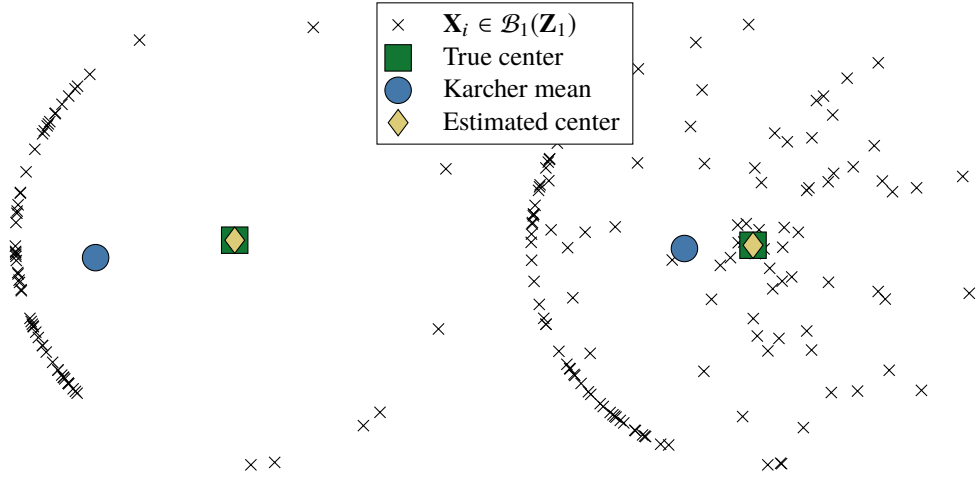}
\caption{Two examples of point sets from Gr$(1,3)$ on the unit ball, $\mathcal{B}_1(\*Z_1)$, sampled with nonuniform density on the boundary, embedded into $\Real^2$ by multidimensional scaling. Points from $\mathcal{B}_1(\*Z_1)$ are indicated with x's, the true center is the green square, the Karcher mean is the blue circle, and the estimated GMEB center is the yellow diamond. }
\label{fig:4d}
\end{figure*}

For $i=1,\ldots,M$, the basis for $\*X_i$ can be completed to a $p_i$ dimensional subspace by taking the span of $X_i$ and $p_i-k_0$ random dimensions. If the $p_i-k_0$ random dimensions are mutually orthogonal for $i=1,\ldots, M$, then the optimal rank for the GMEB is still $k^* = k_0$. Figure~\ref{fig:4d} shows $2$-dimensional embeddings via multidimensional scaling of data sets on Gr$(1,3)$ that have been generated as a unit ball with higher sampling density along a random arc. The yellow diamond indicates the center of the GMEB (computed via the proposed method) and the blue circle marks the Karcher mean of each data collection.

It should be noted that using either data model the point at the center of $\mathcal{B}_{\epsilon_1}(\*Z_1)$ is only the ground-truth center of the minimum enclosing ball of the data collection, $\*U(k^*),$ if the points have been sampled with a high enough density from the surface of the ball. The minimum number uniformly distributed points needed grows with the ambient dimension, $n,$ so in high dimensional spaces the number of points, $M,$ needed to create a ground-truth center may become prohibitively large.  The experimental data can be generated exclusively from the boundary of the balls or interior points can be added.\footnote{Matlab code for the data generation procedures, algorithms, and the numerical experiments are available at \url{https://sites.google.com/site/nicolasgillis/code}.}

%% SECTION 7
\section{Numerical experiments}
\label{sec:numerical}
The experiments in this section are meant to illustrate three properties of the proposed GMEB algorithm and associated order-selection rule. First, we demonstrate the speed and accuracy of the proposed method for estimating the center of the GMEB.  Second, we demonstrate that a warm-start on Gr$(k+1,n)$ using the optimal solution from Gr$(k,n)$ can reduce the number of iterations required for the algorithm to converge.  And finally, we compare results of the proposed order-selection rule and the rule of~\cite{santamaria2016order} in a variety of scenarios to gain intuition about when and how they differ.
\subsection{Experiment 1: Accuracy of the GMEB}
\label{subsec:accuracy}
\begin{figure*}[!t]
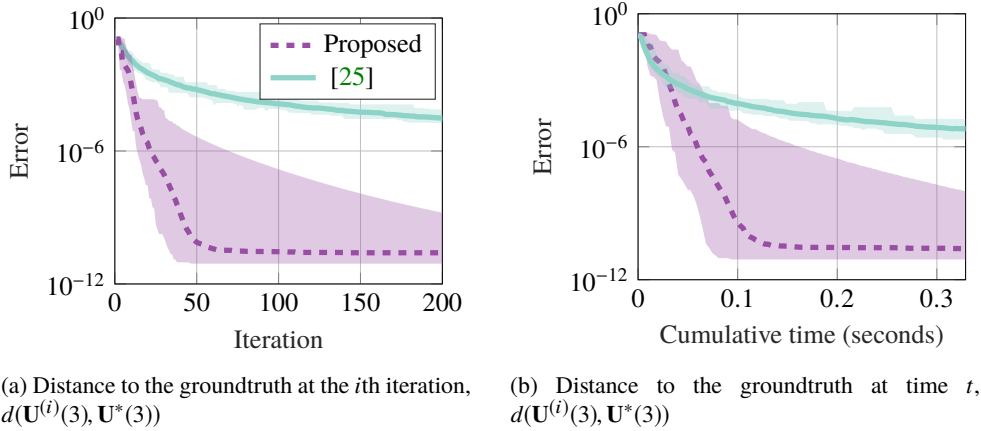

\begin{subfigure}[t]{0.48\linewidth}
\centering
\input{gmeb_fig05a.tex}
\caption{\label{fig:ca_error1}Distance to the groundtruth at the $i$th iteration, $d(\*U^{(i)}(3),\*U^*(3))$}
\end{subfigure}
\hfill
\begin{subfigure}[t]{0.48\linewidth}%
\centering
\input{gmeb_fig05b.tex}
\caption{\label{fig:ca_errorVStime1}Distance to the groundtruth at time $t$, $d(\*U^{(i)}(3),\*U^*(3))$}
\end{subfigure}%
\caption{Median distance to the groundtruth and cumulative time for the GMEB on Gr$(3,10)$ of data generated with the asymmetrical nested ball model from Section~\ref{subsec:nested} over $100$ Monte Carlo trials. The data consists of $100$ points in Gr$(3,10).$ The proposed method is indicated by the dashed purple line and the method of Renard \etal~\cite{renard2018grassmannian} is represented by the solid turquoise line. The shaded regions span the extreme values.}%
\label{fig:center_accuracy1}
\end{figure*}
To test the accuracy and efficiency of the proposed dual subgradient approach, data sets are generated according to the each of two data models from Section~\ref{sec:data}. For each data collection, the GMEB center is approximated using the proposed method and the algorithm of Renard \etal~\cite{renard2018grassmannian}, and the residual error is measured as the between the approximate centers and the true centers. For the first data set, $M=100$ points are sampled from Gr$(3,10)$ using the asymmetrical nested ball model in Section~\ref{subsec:nested} with neither of the proposed generalizations. That is, $k_0 = k_1 = k_2=3$ so that all points are sampled from the same Grassmann manifold. $M_1 = 70$ of the points come from the boundary of $\mathcal{B}_1(\*Z_1)$ and $M_2 = 30$ from the boundary of $\mathcal{B}_{0.125}(\*Z_2)$.  No points are sampled from the interior of either ball. Both algorithms are initialized using the extrinsic mean of the data~\cite{marrinan2014,rentmeesters2010efficient}, that is, $\bm{\lambda}^{(0)} = [ \nicefrac{1}{100},\nicefrac{1}{100}, \ldots, \nicefrac{1}{100}]^T,$ and $\*U^{(0)}(3)$ is the dominant $3$-dimensional eigenspace of $\sum_{i=1}^{100} \lambda^{(0)}_i X_i^{} X_i^T$. The groundtruth center is $\*U^*(3) = \*Z_1.$

Figure~\ref{fig:ca_error1} shows the median distance to the groundtruth over $100$ Monte Carlo trials between the iterate with the lowest primal cost and the ground-truth center. Figure~\ref{fig:ca_errorVStime1} shows the same median distance to the groundtruth relative to cumulative computation time for each algorithm. In both plots the proposed method is indicated by the dashed purple line and the method of~\cite{renard2018grassmannian} is represented by the solid turquoise line. The shaded regions denote the complete range of values across all trials.  This is a setting in which all data points live on a single Grassmann manifold.  Therefore the point-to-set distances reduce to the traditional Grassmannian distances and the technique of~\cite{renard2018grassmannian} is equivalent to that of~\cite{arnaudon2013approximating}. 

The proposed method clearly outperforms the existing technique in terms of accuracy relative to both iterations and computation time for this collection of data. However, the cumulative computation time is affected by many of the parameters in the experimental setup. Let $P = \max_i\{\textrm{dim}(\*X_i)\}.$ For the technique of~\cite{renard2018grassmannian,arnaudon2013approximating}, the per iteration complexity is $\mathcal{O}\left(MP(nk+ k^2)\right)$ due to the $M$ matrix products and subsequent thin SVDs. The proposed method computes these same $M$ products and SVDs, but must additionally compute the compact SVD of a matrix of size $n \times MP$ in order to get the updated center. Assuming that $n \leq MP$ (as it is in all the experiments), the complexity of the proposed algorithm is then $\mathcal{O}\left(MP(nk+k^2 + n^2)\right).$ There are an additional $M$ SVDs for each back-tracking step taken, but those steps are infrequent and thus dominated by the other terms. From these complexities we can see that an increase in the ambient dimension, $n,$ number of subspaces, $M$, or subspace dimension, $P,$ would all lead to a relative decrease in the efficiency of the proposed method.

\begin{figure*}[!t]
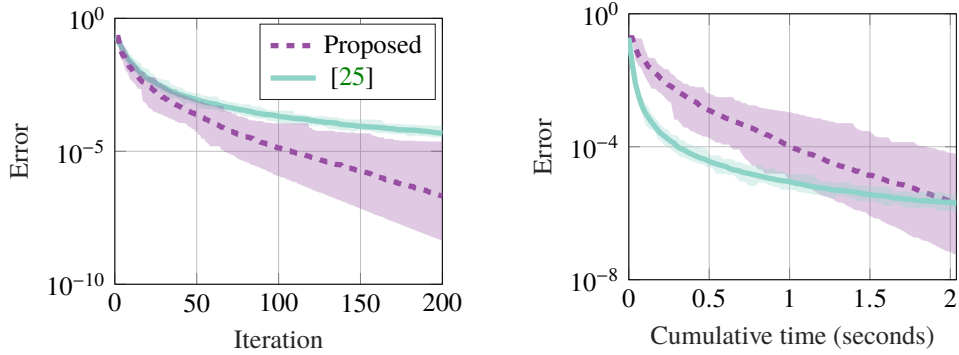

\begin{subfigure}[t]{0.48\linewidth}
\centering
\input{gmeb_fig06a.tex}
\caption{\label{fig:ca_error2}Distance to the groundtruth at the $i$th iteration, $d(\*U^{(i)}(3),\*U^*(3))$}
\end{subfigure}\hfill
\begin{subfigure}[t]{0.48\linewidth}
\centering
\input{gmeb_fig06b.tex}
\caption{\label{fig:ca_errorVStime2}Distance to the groundtruth at time $t$, $d(\*U^{(i)}(3),\*U^*(3))$}
\end{subfigure}%
\caption{Median distance to the groundtruth and cumulative computation time for the GMEB on Gr$(3,15)$ of data generated with the nonuniform sampling model from Section~\ref{subsec:nonuni} over $100$ Monte Carlo trials. The data consists of $300$ points in $\coprod_{p \in \mathcal{P}}{\textrm{Gr}(p,15)}$ for $\mathcal{P} = \{3,4,5,6\}.$ The proposed method is indicated by the dashed purple line and the method of Renard \etal~\cite{renard2018grassmannian} is represented by the solid turquoise line. The shaded regions span the extreme values.}
\label{fig:center_accuracy2}
\end{figure*}

In the second example we employ the data model from Section~\ref{subsec:nonuni}, with the inclusion of interior points and the generalization that the data points come from a disjoint union of Grassmannians, that is, they are subspaces of differing dimensions. Initially, $M_1=100$ points are sampled from the boundary of $\mathcal{B}_1(\*Z_1)$ on Gr$(3,15)$. An additional $M_2=100$ points are selected from an arc on the boundary of the ball between two randomly selected points. Finally $M_3=100$ points are selected uniformly at random from the interior of the ball. Each of the $M = 300$ points is then completed to a basis for a $p_i$-dimensional subspace where $p_i$ is randomly selected from the set $\mathcal{P} = \{3,4,5,6\}$. Both algorithms are again initialized using the extrinsic mean of the data on Gr$(3,15)$ where $\bm{\lambda}^{(0)} = [ \nicefrac{1}{300},\nicefrac{1}{300}, \ldots, \nicefrac{1}{300}]^T,$ and $\*U^{(0)}(3)$ is the dominant $3$-dimensional eigenspace of $\sum_{i=1}^{300} \lambda^{(0)}_i X_i^{} X_i^T$. Figure~\ref{fig:ca_error2} shows the median distance to the groundtruth over $100$ Monte Carlo trials between the iterate with the lowest primal cost and the ground-truth center, while Figure~\ref{fig:ca_errorVStime2} shows the median error relative to cumulative computation time. The proposed method is indicated by the dashed purple line and the method of Renard \etal~\cite{renard2018grassmannian} is represented by the solid turquoise line. The shaded regions span the extreme values. The groundtruth center is $\*U^*(3) = \*Z_1.$

As shown in Figure~\ref{fig:ca_error2}, the proposed method achieves a higher accuracy in fewer iterations than~\cite{renard2018grassmannian}. However, the greater complexity of the proposed method means that the primal algorithm initially achieves a lower error, as shown in Figure~\ref{fig:ca_errorVStime2}. The increased number of points in the data set and specifically in the support of the GMEB lead to a slower overall convergence for the proposed algorithm. This reduced efficiency would grow with the size of the data, however the subgradient technique is consistently achieves lower overall error given enough time. Moreover, the proposed method provides duality-gap optimality guarantees. 

One direction for future work is to combine the two methods to get the best of both worlds; fast initial estimates of the center and high accuracy solutions over time. Using $\*U^{(t)}(k)$ computed via $t$ iterations of~\cite{renard2018grassmannian} as an estimate of the center, we can find dual-feasible variables that are non-zero only for points in the support set of the enclosing ball centered at $\*U^{(t)}(k).$  For example, let $\mathcal{I} = \{i : d_{\textrm{Gr}(k,n)}(\*U^{(t)}(k),\*X_i) = \max_i d_{\textrm{Gr}(k,n)}(\*U^{(t)}(k),\*X_i)\}.$ Then let $\lambda_i^{(0)} = \nicefrac{1}{|\mathcal{I}|}$ for $i \in \mathcal{I}$ and $\lambda_i^{(0)} = 0 $ otherwise, and proceed with the subgradient algorithm from this warm-start. An alternative initialization strategy is proposed in Section~\ref{subsec:warm_start}.

\subsection{Experiment 2: Faster convergence by initializing with previous solutions}
\label{subsec:warm_start}
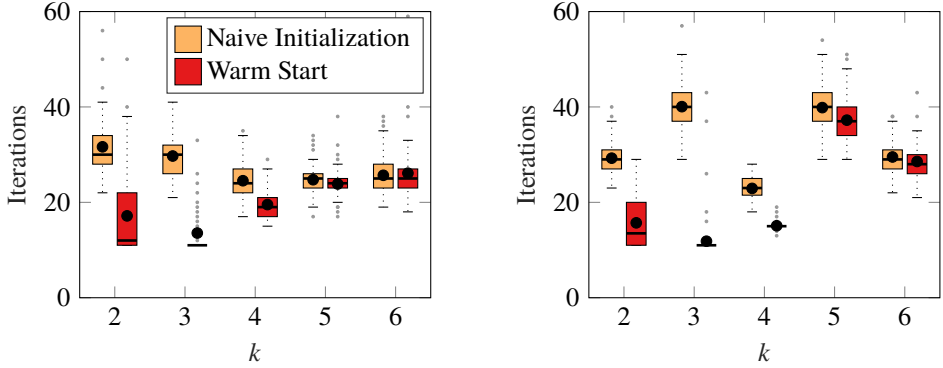
\begin{figure*}[!t]
\begin{subfigure}[t]{0.48\linewidth}
\centering
%% Figure 7a
\definecolor{mycolor1}{rgb}{0.00000,0.44700,0.74100}%
\definecolor{mycolor2}{rgb}{0.99216,0.70588,0.38431}%
\definecolor{mycolor3}{rgb}{0.85000,0.32500,0.09800}%
\definecolor{mycolor4}{rgb}{0.92900,0.69400,0.12500}%
\definecolor{mycolor5}{rgb}{0.49400,0.18400,0.55600}%
\definecolor{mycolor6}{rgb}{0.46600,0.67400,0.18800}%
\definecolor{mycolor7}{rgb}{0.30100,0.74500,0.93300}%
\definecolor{mycolor8}{rgb}{0.63500,0.07800,0.18400}%
\definecolor{mycolor9}{rgb}{0.89412,0.10196,0.10980}%

\begin{tikzpicture}
\begin{axis}[%
width=\linewidth,
at={(0.614in,0.52in)},
xmin=0.5,
xmax=5.5,
xtick={1,2,3,4,5},
xticklabels={{2},{3},{4},{5},{6}},
xlabel style={font=\color{white!15!black}},
xlabel={$k$},
ymin=0,
ymax=60,
ylabel style={font=\color{white!15!black}},
ylabel={Iterations},
axis background/.style={fill=white},
legend style={legend cell align=left, align=left, draw=white!15!black}
]
\addplot [color=mycolor1, draw=none, mark size=0.5pt, mark=*, mark options={solid, fill=white!60!black, white!60!black}, forget plot]
  table[row sep=crcr]{%
0.825000000000003	44\\
0.825000000000003	50\\
0.825000000000003	56\\
};
\addplot [color=black, dotted, forget plot]
  table[row sep=crcr]{%
0.825000000000003	41\\
0.825000000000003	22\\
};
\addplot [color=black, forget plot]
  table[row sep=crcr]{%
0.755000000000003	41\\
0.895000000000003	41\\
};
\addplot [color=black, forget plot]
  table[row sep=crcr]{%
0.754999999999999	22\\
0.895	22\\
};
\draw[fill=mycolor2, draw=black] (axis cs:0.685,28) rectangle (axis cs:0.965,34);
\addplot [color=black, line width=1.0pt, forget plot]
  table[row sep=crcr]{%
0.684999999999999	30\\
0.965	30\\
};
\addplot [color=mycolor3, draw=none, mark=*, mark options={solid, fill=mycolor2, black}, forget plot]
  table[row sep=crcr]{%
0.824999999999999	31.64\\
};
\addplot [color=black, dotted, forget plot]
  table[row sep=crcr]{%
1.825	41\\
1.825	21\\
};
\addplot [color=black, forget plot]
  table[row sep=crcr]{%
1.755	41\\
1.895	41\\
};
\addplot [color=black, forget plot]
  table[row sep=crcr]{%
1.755	21\\
1.895	21\\
};
\draw[fill=mycolor2, draw=black] (axis cs:1.685,26) rectangle (axis cs:1.965,32);
\addplot [color=black, line width=1.0pt, forget plot]
  table[row sep=crcr]{%
1.685	30\\
1.965	30\\
};
\addplot [color=mycolor4, draw=none, mark=*, mark options={solid, fill=mycolor2, black}, forget plot]
  table[row sep=crcr]{%
1.825	29.71\\
};
\addplot [color=mycolor5, draw=none, mark size=0.5pt, mark=*, mark options={solid, fill=white!60!black, white!60!black}, forget plot]
  table[row sep=crcr]{%
2.825	35\\
};
\addplot [color=black, dotted, forget plot]
  table[row sep=crcr]{%
2.825	34\\
2.825	17\\
};
\addplot [color=black, forget plot]
  table[row sep=crcr]{%
2.755	34\\
2.895	34\\
};
\addplot [color=black, forget plot]
  table[row sep=crcr]{%
2.755	17\\
2.895	17\\
};
\draw[fill=mycolor2, draw=black] (axis cs:2.685,22) rectangle (axis cs:2.965,27);
\addplot [color=black, line width=1.0pt, forget plot]
  table[row sep=crcr]{%
2.685	24\\
2.965	24\\
};
\addplot [color=mycolor6, draw=none, mark=*, mark options={solid, fill=mycolor2, black}, forget plot]
  table[row sep=crcr]{%
2.825	24.53\\
};
\addplot [color=mycolor7, draw=none, mark size=0.5pt, mark=*, mark options={solid, fill=white!60!black, white!60!black}, forget plot]
  table[row sep=crcr]{%
3.825	31\\
3.825	32\\
3.825	33\\
3.825	34\\
};
\addplot [color=mycolor8, draw=none, mark size=0.5pt, mark=*, mark options={solid, fill=white!60!black, white!60!black}, forget plot]
  table[row sep=crcr]{%
3.825	17\\
};
\addplot [color=black, dotted, forget plot]
  table[row sep=crcr]{%
3.825	29\\
3.825	19\\
};
\addplot [color=black, forget plot]
  table[row sep=crcr]{%
3.755	29\\
3.895	29\\
};
\addplot [color=black, forget plot]
  table[row sep=crcr]{%
3.755	19\\
3.895	19\\
};
\draw[fill=mycolor2, draw=black] (axis cs:3.685,23) rectangle (axis cs:3.965,26);
\addplot [color=black, line width=1.0pt, forget plot]
  table[row sep=crcr]{%
3.685	25\\
3.965	25\\
};
\addplot [color=mycolor1, draw=none, mark=*, mark options={solid, fill=mycolor2, black}, forget plot]
  table[row sep=crcr]{%
3.825	24.74\\
};
\addplot [color=mycolor3, draw=none, mark size=0.5pt, mark=*, mark options={solid, fill=white!60!black, white!60!black}, forget plot]
  table[row sep=crcr]{%
4.825	36\\
4.825	37\\
4.825	38\\
4.825	44\\
};
\addplot [color=black, dotted, forget plot]
  table[row sep=crcr]{%
4.825	35\\
4.825	19\\
};
\addplot [color=black, forget plot]
  table[row sep=crcr]{%
4.755	35\\
4.895	35\\
};
\addplot [color=black, forget plot]
  table[row sep=crcr]{%
4.755	19\\
4.895	19\\
};
\draw[fill=mycolor2, draw=black] (axis cs:4.685,23) rectangle (axis cs:4.965,28);
\addplot [color=black, line width=1.0pt, forget plot]
  table[row sep=crcr]{%
4.685	25\\
4.965	25\\
};
\addplot [color=mycolor4, draw=none, mark=*, mark options={solid, fill=mycolor2, black}, forget plot]
  table[row sep=crcr]{%
4.825	25.67\\
};

\addplot [color=mycolor5, draw=none, mark size=0.5pt, mark=*, mark options={solid, fill=white!60!black, white!60!black}, forget plot]
  table[row sep=crcr]{%
1.175	40\\
1.175	50\\
};
\addplot [color=black, dotted, forget plot]
  table[row sep=crcr]{%
1.175	38\\
1.175	11\\
};
\addplot [color=black, forget plot]
  table[row sep=crcr]{%
1.105	38\\
1.245	38\\
};
\addplot [color=black, forget plot]
  table[row sep=crcr]{%
1.105	11\\
1.245	11\\
};
\draw[fill=mycolor9, draw=black] (axis cs:1.035,11) rectangle (axis cs:1.315,22);
\addplot [color=black, line width=1.0pt, forget plot]
  table[row sep=crcr]{%
1.035	12\\
1.315	12\\
};
\addplot [color=mycolor6, draw=none, mark=*, mark options={solid, fill=mycolor9, black}, forget plot]
  table[row sep=crcr]{%
1.175	17.16\\
};
\addplot [color=mycolor7, draw=none, mark size=0.5pt, mark=*, mark options={solid, fill=white!60!black, white!60!black}, forget plot]
  table[row sep=crcr]{%
2.175	12\\
2.175	13\\
2.175	14\\
2.175	15\\
2.175	16\\
2.175	17\\
2.175	18\\
2.175	19\\
2.175	20\\
2.175	22\\
2.175	24\\
2.175	26\\
2.175	33\\
};
\addplot [color=black, dotted, forget plot]
  table[row sep=crcr]{%
2.175	11\\
2.175	11\\
};
\addplot [color=black, forget plot]
  table[row sep=crcr]{%
2.105	11\\
2.245	11\\
};
\addplot [color=black, forget plot]
  table[row sep=crcr]{%
2.105	11\\
2.245	11\\
};
\addplot [color=black, line width=1.0pt, forget plot]
  table[row sep=crcr]{%
2.035	11\\
2.315	11\\
};
\addplot [color=mycolor8, draw=none, mark=*, mark options={solid, fill=mycolor9, black}, forget plot]
  table[row sep=crcr]{%
2.175	13.56\\
};
\addplot [color=mycolor1, draw=none, mark size=0.5pt, mark=*, mark options={solid, fill=white!60!black, white!60!black}, forget plot]
  table[row sep=crcr]{%
3.175	29\\
};
\addplot [color=black, dotted, forget plot]
  table[row sep=crcr]{%
3.175	27\\
3.175	15\\
};
\addplot [color=black, forget plot]
  table[row sep=crcr]{%
3.105	27\\
3.245	27\\
};
\addplot [color=black, forget plot]
  table[row sep=crcr]{%
3.105	15\\
3.245	15\\
};
\draw[fill=mycolor9, draw=black] (axis cs:3.035,17) rectangle (axis cs:3.315,21);
\addplot [color=black, line width=1.0pt, forget plot]
  table[row sep=crcr]{%
3.035	19\\
3.315	19\\
};
\addplot [color=mycolor3, draw=none, mark=*, mark options={solid, fill=mycolor9, black}, forget plot]
  table[row sep=crcr]{%
3.175	19.53\\
};
\addplot [color=mycolor4, draw=none, mark size=0.5pt, mark=*, mark options={solid, fill=white!60!black, white!60!black}, forget plot]
  table[row sep=crcr]{%
4.175	29\\
4.175	30\\
4.175	32\\
4.175	38\\
};
\addplot [color=mycolor5, draw=none, mark size=0.5pt, mark=*, mark options={solid, fill=white!60!black, white!60!black}, forget plot]
  table[row sep=crcr]{%
4.175	17\\
4.175	18\\
4.175	19\\
4.175	19\\
};
\addplot [color=black, dotted, forget plot]
  table[row sep=crcr]{%
4.175	28\\
4.175	20\\
};
\addplot [color=black, forget plot]
  table[row sep=crcr]{%
4.105	28\\
4.245	28\\
};
\addplot [color=black, forget plot]
  table[row sep=crcr]{%
4.105	20\\
4.245	20\\
};
\draw[fill=mycolor9, draw=black] (axis cs:4.035,23) rectangle (axis cs:4.315,25);
\addplot [color=black, line width=1.0pt, forget plot]
  table[row sep=crcr]{%
4.035	24\\
4.315	24\\
};
\addplot [color=mycolor6, draw=none, mark=*, mark options={solid, fill=mycolor9, black}, forget plot]
  table[row sep=crcr]{%
4.175	23.85\\
};
\addplot [color=mycolor7, draw=none, mark size=0.5pt, mark=*, mark options={solid, fill=white!60!black, white!60!black}, forget plot]
  table[row sep=crcr]{%
5.175	38\\
5.175	40\\
5.175	48\\
5.175	59\\
};
\addplot [color=black, dotted, forget plot]
  table[row sep=crcr]{%
5.175	33\\
5.175	18\\
};
\addplot [color=black, forget plot]
  table[row sep=crcr]{%
5.105	33\\
5.245	33\\
};
\addplot [color=black, forget plot]
  table[row sep=crcr]{%
5.105	18\\
5.245	18\\
};
\draw[fill=mycolor9, draw=black] (axis cs:5.035,23) rectangle (axis cs:5.315,27);
\addplot [color=black, line width=1.0pt, forget plot]
  table[row sep=crcr]{%
5.035	25\\
5.315	25\\
};
\addplot [color=mycolor8, draw=none, mark=*, mark options={solid, fill=mycolor9, black}, forget plot]
  table[row sep=crcr]{%
5.175	26.11\\
};

\addlegendimage{only marks, mark=square*, mark options={}, mark size=5pt, draw=black, fill=mycolor2}
\addlegendentry{Naive Initialization}
\addlegendimage{only marks, mark=square*, mark options={}, mark size=5pt, draw=black, fill=mycolor9}
\addlegendentry{Warm Start}

\end{axis}
\end{tikzpicture}%
\caption{\label{fig:warm_iter1} Results from $100$ trials with the asymmetrical nested ball model where $k^* = 4$ and $M = 50$ points sampled from Gr$(p_i,10)$ with $p_i \in \{4,5,6\}$.}
\end{subfigure}\hfill
\begin{subfigure}[t]{0.48\linewidth}
\centering
%% Figure 7b
\definecolor{mycolor1}{rgb}{0.00000,0.44700,0.74100}%
\definecolor{mycolor2}{rgb}{0.99216,0.70588,0.38431}%
\definecolor{mycolor3}{rgb}{0.85000,0.32500,0.09800}%
\definecolor{mycolor4}{rgb}{0.92900,0.69400,0.12500}%
\definecolor{mycolor5}{rgb}{0.49400,0.18400,0.55600}%
\definecolor{mycolor6}{rgb}{0.46600,0.67400,0.18800}%
\definecolor{mycolor7}{rgb}{0.30100,0.74500,0.93300}%
\definecolor{mycolor8}{rgb}{0.63500,0.07800,0.18400}%
\definecolor{mycolor9}{rgb}{0.89412,0.10196,0.10980}%

\begin{tikzpicture}
\begin{axis}[%
width=\linewidth,
at={(0.614in,0.52in)},
xmin=0.5,
xmax=5.5,
xtick={1,2,3,4,5},
xticklabels={{2},{3},{4},{5},{6}},
xlabel style={font=\color{white!15!black}},
xlabel={$k$},
ymin=0,
ymax=60,
ylabel style={font=\color{white!15!black}},
ylabel={Iterations},
axis background/.style={fill=white},
legend style={legend cell align=left, align=left, draw=white!15!black}
]
\addplot [color=mycolor1, draw=none, mark size=0.5pt, mark=*, mark options={solid, fill=white!60!black, white!60!black}, forget plot]
  table[row sep=crcr]{%
0.825000000000003	38\\
0.825000000000003	40\\
};
\addplot [color=black, dotted, forget plot]
  table[row sep=crcr]{%
0.825000000000003	37\\
0.825000000000003	23\\
};
\addplot [color=black, forget plot]
  table[row sep=crcr]{%
0.755000000000003	37\\
0.895000000000003	37\\
};
\addplot [color=black, forget plot]
  table[row sep=crcr]{%
0.754999999999999	23\\
0.895	23\\
};
\draw[fill=mycolor2, draw=black] (axis cs:0.685,27) rectangle (axis cs:0.965,31);
\addplot [color=black, line width=1.0pt, forget plot]
  table[row sep=crcr]{%
0.684999999999999	29\\
0.965	29\\
};
\addplot [color=mycolor3, draw=none, mark=*, mark options={solid, fill=mycolor2, black}, forget plot]
  table[row sep=crcr]{%
0.824999999999999	29.26\\
};
\addplot [color=mycolor4, draw=none, mark size=0.5pt, mark=*, mark options={solid, fill=white!60!black, white!60!black}, forget plot]
  table[row sep=crcr]{%
1.825	57\\
};
\addplot [color=black, dotted, forget plot]
  table[row sep=crcr]{%
1.825	51\\
1.825	29\\
};
\addplot [color=black, forget plot]
  table[row sep=crcr]{%
1.755	51\\
1.895	51\\
};
\addplot [color=black, forget plot]
  table[row sep=crcr]{%
1.755	29\\
1.895	29\\
};
\draw[fill=mycolor2, draw=black] (axis cs:1.685,37) rectangle (axis cs:1.965,43);
\addplot [color=black, line width=1.0pt, forget plot]
  table[row sep=crcr]{%
1.685	40\\
1.965	40\\
};
\addplot [color=mycolor5, draw=none, mark=*, mark options={solid, fill=mycolor2, black}, forget plot]
  table[row sep=crcr]{%
1.825	40.06\\
};
\addplot [color=black, dotted, forget plot]
  table[row sep=crcr]{%
2.825	28\\
2.825	18\\
};
\addplot [color=black, forget plot]
  table[row sep=crcr]{%
2.755	28\\
2.895	28\\
};
\addplot [color=black, forget plot]
  table[row sep=crcr]{%
2.755	18\\
2.895	18\\
};
\draw[fill=mycolor2, draw=black] (axis cs:2.685,21.5) rectangle (axis cs:2.965,25);
\addplot [color=black, line width=1.0pt, forget plot]
  table[row sep=crcr]{%
2.685	23\\
2.965	23\\
};
\addplot [color=mycolor6, draw=none, mark=*, mark options={solid, fill=mycolor2, black}, forget plot]
  table[row sep=crcr]{%
2.825	22.92\\
};
\addplot [color=mycolor7, draw=none, mark size=0.5pt, mark=*, mark options={solid, fill=white!60!black, white!60!black}, forget plot]
  table[row sep=crcr]{%
3.825	54\\
};
\addplot [color=black, dotted, forget plot]
  table[row sep=crcr]{%
3.825	51\\
3.825	29\\
};
\addplot [color=black, forget plot]
  table[row sep=crcr]{%
3.755	51\\
3.895	51\\
};
\addplot [color=black, forget plot]
  table[row sep=crcr]{%
3.755	29\\
3.895	29\\
};
\draw[fill=mycolor2, draw=black] (axis cs:3.685,37) rectangle (axis cs:3.965,43);
\addplot [color=black, line width=1.0pt, forget plot]
  table[row sep=crcr]{%
3.685	40\\
3.965	40\\
};
\addplot [color=mycolor8, draw=none, mark=*, mark options={solid, fill=mycolor2, black}, forget plot]
  table[row sep=crcr]{%
3.825	39.86\\
};
\addplot [color=mycolor1, draw=none, mark size=0.5pt, mark=*, mark options={solid, fill=white!60!black, white!60!black}, forget plot]
  table[row sep=crcr]{%
4.825	38\\
4.825	38\\
};
\addplot [color=black, dotted, forget plot]
  table[row sep=crcr]{%
4.825	37\\
4.825	22\\
};
\addplot [color=black, forget plot]
  table[row sep=crcr]{%
4.755	37\\
4.895	37\\
};
\addplot [color=black, forget plot]
  table[row sep=crcr]{%
4.755	22\\
4.895	22\\
};
\draw[fill=mycolor2, draw=black] (axis cs:4.685,27) rectangle (axis cs:4.965,31);
\addplot [color=black, line width=1.0pt, forget plot]
  table[row sep=crcr]{%
4.685	29\\
4.965	29\\
};
\addplot [color=mycolor3, draw=none, mark=*, mark options={solid, fill=mycolor2, black}, forget plot]
  table[row sep=crcr]{%
4.825	29.51\\
};

\addplot [color=black, dotted, forget plot]
  table[row sep=crcr]{%
1.175	29\\
1.175	11\\
};
\addplot [color=black, forget plot]
  table[row sep=crcr]{%
1.105	29\\
1.245	29\\
};
\addplot [color=black, forget plot]
  table[row sep=crcr]{%
1.105	11\\
1.245	11\\
};
\draw[fill=mycolor9, draw=black] (axis cs:1.035,11) rectangle (axis cs:1.315,20);
\addplot [color=black, line width=1.0pt, forget plot]
  table[row sep=crcr]{%
1.035	13.5\\
1.315	13.5\\
};
\addplot [color=mycolor4, draw=none, mark=*, mark options={solid, fill=mycolor9, black}, forget plot]
  table[row sep=crcr]{%
1.175	15.7\\
};
\addplot [color=mycolor5, draw=none, mark size=0.5pt, mark=*, mark options={solid, fill=white!60!black, white!60!black}, forget plot]
  table[row sep=crcr]{%
2.175	16\\
2.175	18\\
2.175	26\\
2.175	37\\
2.175	43\\
};
\addplot [color=black, dotted, forget plot]
  table[row sep=crcr]{%
2.175	11\\
2.175	11\\
};
\addplot [color=black, forget plot]
  table[row sep=crcr]{%
2.105	11\\
2.245	11\\
};
\addplot [color=black, forget plot]
  table[row sep=crcr]{%
2.105	11\\
2.245	11\\
};
\addplot [color=black, line width=1.0pt, forget plot]
  table[row sep=crcr]{%
2.035	11\\
2.315	11\\
};
\addplot [color=mycolor6, draw=none, mark=*, mark options={solid, fill=mycolor9, black}, forget plot]
  table[row sep=crcr]{%
2.175	11.85\\
};
\addplot [color=mycolor7, draw=none, mark size=0.5pt, mark=*, mark options={solid, fill=white!60!black, white!60!black}, forget plot]
  table[row sep=crcr]{%
3.175	16\\
3.175	17\\
3.175	18\\
3.175	19\\
};
\addplot [color=mycolor8, draw=none, mark size=0.5pt, mark=*, mark options={solid, fill=white!60!black, white!60!black}, forget plot]
  table[row sep=crcr]{%
3.175	13\\
3.175	14\\
3.175	14\\
};
\addplot [color=black, dotted, forget plot]
  table[row sep=crcr]{%
3.175	15\\
3.175	15\\
};
\addplot [color=black, forget plot]
  table[row sep=crcr]{%
3.105	15\\
3.245	15\\
};
\addplot [color=black, forget plot]
  table[row sep=crcr]{%
3.105	15\\
3.245	15\\
};
\addplot [color=black, line width=1.0pt, forget plot]
  table[row sep=crcr]{%
3.035	15\\
3.315	15\\
};
\addplot [color=mycolor1, draw=none, mark=*, mark options={solid, fill=mycolor9, black}, forget plot]
  table[row sep=crcr]{%
3.175	15.09\\
};
\addplot [color=mycolor3, draw=none, mark size=0.5pt, mark=*, mark options={solid, fill=white!60!black, white!60!black}, forget plot]
  table[row sep=crcr]{%
4.175	50\\
4.175	51\\
};
\addplot [color=black, dotted, forget plot]
  table[row sep=crcr]{%
4.175	48\\
4.175	29\\
};
\addplot [color=black, forget plot]
  table[row sep=crcr]{%
4.105	48\\
4.245	48\\
};
\addplot [color=black, forget plot]
  table[row sep=crcr]{%
4.105	29\\
4.245	29\\
};
\draw[fill=mycolor9, draw=black] (axis cs:4.035,34) rectangle (axis cs:4.315,40);
\addplot [color=black, line width=1.0pt, forget plot]
  table[row sep=crcr]{%
4.035	37\\
4.315	37\\
};
\addplot [color=mycolor4, draw=none, mark=*, mark options={solid, fill=mycolor9, black}, forget plot]
  table[row sep=crcr]{%
4.175	37.23\\
};
\addplot [color=mycolor5, draw=none, mark size=0.5pt, mark=*, mark options={solid, fill=white!60!black, white!60!black}, forget plot]
  table[row sep=crcr]{%
5.175	38\\
5.175	43\\
};
\addplot [color=black, dotted, forget plot]
  table[row sep=crcr]{%
5.175	35\\
5.175	21\\
};
\addplot [color=black, forget plot]
  table[row sep=crcr]{%
5.105	35\\
5.245	35\\
};
\addplot [color=black, forget plot]
  table[row sep=crcr]{%
5.105	21\\
5.245	21\\
};
\draw[fill=mycolor9, draw=black] (axis cs:5.035,26) rectangle (axis cs:5.315,30);
\addplot [color=black, line width=1.0pt, forget plot]
  table[row sep=crcr]{%
5.035	28\\
5.315	28\\
};
\addplot [color=mycolor6, draw=none, mark=*, mark options={solid, fill=mycolor9, black}, forget plot]
  table[row sep=crcr]{%
5.175	28.57\\
};

\end{axis}
\end{tikzpicture}%
\caption{\label{fig:warm_iter2} Results from $100$ trials with the nonuniform sampling model where $k^* = 4$ and $M = 300$ points sampled from Gr$(p_i,10)$ with $p_i \in \{4,5,6\}$.}
\end{subfigure}
\caption{\label{fig:warm}Number of iterations needed for the proposed subgradient algorithm to reach a stationary point using a naive initialization, $\bm{\lambda}^{(0)}(k+1) = [ \nicefrac{1}{M},\nicefrac{1}{M}, \ldots, \nicefrac{1}{M}]^T$ (light orange), and a warm start,  $\bm{\lambda}^{(0)}(k+1) =\bm{\lambda}^{*}(k)$ (red) for two data sets. }
\end{figure*}
To apply the order selection criteria in Section~\ref{sec:ord_select}, the GMEB center must be computed for $k=1,\ldots,\max_i\{\textrm{dim}(\*X_i)\}.$ The example in Section~\ref{subsec:not_nested} demonstrates that the subspace at the center of the minimum enclosing ball cannot be built in a greedy fashion, because the center $\*U^*(k-1) \in \textrm{Gr}(k-1,n)$ is not in general a subspace of the center $\*U^*(k) \in \textrm{Gr}(k,n)$. However, the solutions are often \textit{nearly} nested.  As a result, the vector, $\bm{\lambda}^*(k-1),$ that provides the optimal value of the dual objective function for the problem on Gr$(k-1,n)$ can offer a good initialization for the dual subgradient algorithm used to find the GMEB center on Gr$(k,n),$ significantly reducing the total computation time needed to identify the optimal dimension, $k^*$.  By way of comparison, simple initializations of $\bm{\lambda}^{(0)}(k)$ would be to randomly select the dual variables or to set all of the dual variables equal so that $\bm{\lambda}^{(0)}(k)=[\nicefrac{1}{M},\ldots, \nicefrac{1}{M}]^T.$ For these experiments the latter strategy is chosen. The initial iterate for the primal variable when the dual variables are all equal is then the uniformly weighted extrinsic mean of the data, that is, $\*U^{(0)}(k)$ is the dominant $k$-dimensional eigenspace of $\sum_{i=1}^{M} \lambda^{(0)}_i X_i^{} X_i^T.$  On Gr$(1,n),$ no warm-start initialization is possible because $\bm{\lambda}^{*}(0)$ is undefined, so the algorithm is run using only the naive initialization.  For $k=2,\ldots,\max_i\{\textrm{dim}(\*X_i)\}$ Figure~\ref{fig:warm} illustrates the relative speed-up due to smart initialization by comparing the number of iterations needed to find a stationary point for different choices of the initial dual variable using each of the data models. Both data models are intentionally structured so that the extrinsic mean is not the center of the GMEB on Gr$(k^*,n)$. The naive initialization is indicated by the light orange box-and-whisker plots, while the warm-start is denoted with red. The black dots mark the mean number of iterations and the solid line is the median.

In Figure~\ref{fig:warm_iter1} the data has been generated using the asymmetrical nested ball model with $M=50$ points sampled from Gr$(p_i,10)$ for $p_i \in \{4,5,6\}$ and an optimal dimension of $k^*=4$. The warm start converged in less iterations than the naive initialization in $359$ out of $500$ possible trials. An experiment using data generated by sampling more densely from a randomly selected arc of a unit ball is displayed in Figure~\ref{fig:warm_iter2}.  Here, $M=300$ points were generated on Gr$(p_i,10)$ with $p_i \in \{4,5,6\}$ where $k^*=4$. In $415$ out of $500$ possible trials, the warm start converged in less iterations than the naive initialization. 

\subsection{Experiment 3: Order-selection comparison}
\label{subsec:order_selection}
The previous experiments demonstrated the effectiveness of the proposed approach for computing the subspace at the center of the GMEB in a noise-free scenario.  However the end-goal is to find a central subspace \textit{and} the optimal size to best represent the common dimensions in a collection of data. Adding noise to the subspaces makes it difficult to identify how many common dimensions exist, thus the third experiment compares the ability of the proposed order-selection rule to identify the optimal dimension of the common subspace with that of the technique from Santamaria \etal~\cite{santamaria2016order} as the difficulty of the task varies. 

In many machine learning applications, extracting a low-rank common subspace from data is a pre-processing task and the rank is selected with little care. Heuristic solutions often focus on different methods for locating include the elbow of the scree plot, that is, computing the SVD of the concatenated data sets, finding the  the singular values that represent the significant information, and keeping the dimensions corresponding to these singular values. This can be done with a variety of techniques such as the L-method~\cite{salvador2004determining}, which estimates the elbow as the intersection of the two lines that minimize the root mean-squared error of the projection of the points in the of the scree plot onto the lines, the method of~\cite{zhu2006automatic}, which maximizes the profile log-likelihood under an independence assumption, and even just visually inspecting the scree plot to identify the first significant change in the first derivative~\cite{steyvers2006multidimensional}. To justify the need for a more principled way of selecting a subspace dimension, we additionally compare to the elbow of the scree plot using the L-method, and expect it to provide bad results. In the experiments this technique is denoted ``SVD.''

Figure~\ref{fig:opt_dim2} shows a comparison of order-selection rules for $M=20$ points generated using the asymmetrical nested ball model from Section~\ref{subsec:nested} with both generalizations. The data has $M_1 = 10$ points are sampled uniformly from the boundary of $\mathcal{B}_{1}(\*Z_1) \subset \textrm{Gr}(10,n)$ and $M_2=10$ points are sampled from the boundary of $\mathcal{B}_{.5}(\*Z_2) \subset \textrm{Gr}(15,n).$ Each of the points is then completed to a basis for a point on Gr$(p_i,n)$ for $p_i \in \{10,11, \ldots,20\}$ and $n = 20, 30, \ldots, 200.$ Zero-mean Gaussian noise is added to each basis to create noisy data sets. The signal-to-noise ratio (SNR) of the data is the total power of the signal divided by the total power of the noise.  In order to have the same SNR for each subspace despite differing dimensions, the noise variance per component is scaled by the number of subspace dimensions. Since $X_i$ is an orthonormal basis for $\*X_i,$ the magnitude of each basis vector is $1.$ Thus the total power of signal subspace is $k^*,$ and the SNR is computed as $\textrm{SNR} = 10\log_{10}(\nicefrac{k^*}{\sigma_N^2}),$ where $\sigma_N^2$ is the total variance of the noise. In this example the order of the common subspace is $k^* = 10$ and $\sigma_N^2 = 1.259$ meaning that the data has an SNR of $9$dB.
\begin{figure*}[!t]
\begin{subfigure}[t]{0.5\linewidth}
\centering
%% Figure 8a
\definecolor{mycolor1}{rgb}{0.98431,0.50196,0.44706}%
\definecolor{mycolor2}{rgb}{0.59608,0.30588,0.63922}%
\definecolor{mycolor3}{rgb}{0.99216,0.70588,0.38431}%
\definecolor{mycolor4}{rgb}{0.55294,0.82745,0.78039}%

\begin{tikzpicture}
\begin{axis}[%
width=.95\linewidth,
at={(0.614in,0.52in)},
xmin=20,
xmax=200,
xlabel style={font=\color{white!15!black}},
xlabel={Ambient dimension, $n$},
xtick={20,50,100,150,200},
ymin=0,
ymax=1,
ylabel style={font=\color{white!15!black}},
ylabel={Accuracy},
axis background/.style={fill=white},
reverse legend,
grid=both,
grid style={line width=.1pt, draw=gray!10},
major grid style={line width=.2pt,draw=gray!50},
legend style={legend cell align=left, align=left, at={(0.97,0.3)},anchor=east, draw=white!15!black}
]

% SVD
\addplot [color=mycolor3, dashdotted, line width=2.0pt, mark=x, mark options={solid, mycolor3}]
  table[row sep=crcr]{%
20	0\\
30	0\\
40	0\\
50	0\\
60	0\\
70	0\\
80	0\\
90	0\\
100	0\\
110	0\\
120	0\\
130	0\\
140	0\\
150	0\\
160	0\\
170	0\\
180	0\\
190	0\\
200	0\\
};
\addlegendentry{SVD}

% Hybrid
\addplot [color=mycolor4, densely dotted, line width=2.0pt, mark=square, mark options={solid, mycolor4}]
  table[row sep=crcr]{%
20	0.0200000000000102\\
30	0.780000000000001\\
40	0.789999999999992\\
50	0.939999999999998\\
60	0.930000000000007\\
70	0.949999999999989\\
80	1\\
90	0.990000000000009\\
100	1\\
110	0.990000000000009\\
120	1\\
130	1\\
140	1\\
150	1\\
160	0.990000000000009\\
170	1\\
180	1\\
190	1\\
200	1\\
};
\addlegendentry{Hybrid}

% Santamaria
\addplot [color=mycolor1, line width=2.0pt, mark=o, mark options={solid, mycolor1}]
  table[row sep=crcr]{%
20	0\\
30	0\\
40	0\\
50	0\\
60	0\\
70	0\\
80	0\\
90	0.0800000000000125\\
100	0.129999999999995\\
110	0.330000000000013\\
120	0.47999999999999\\
130	0.789999999999992\\
140	0.909999999999997\\
150	0.960000000000008\\
160	0.97999999999999\\
170	1\\
180	1\\
190	1\\
200	1\\
};
\addlegendentry{\cite{santamaria2016order}}

% Proposed
\addplot [color=mycolor2, dashed, mark=triangle, mark options={solid, mycolor2}, line width=2.0pt]
  table[row sep=crcr]{%
20	0.22999999999999\\
30	0.370000000000005\\
40	0.610000000000014\\
50	0.840000000000003\\
60	0.900000000000006\\
70	0.889999999999986\\
80	0.960000000000008\\
90	0.97999999999999\\
100	0.97999999999999\\
110	0.97999999999999\\
120	0.960000000000008\\
130	0.97999999999999\\
140	0.97999999999999\\
150	0.939999999999998\\
160	0.990000000000009\\
170	0.990000000000009\\
180	0.97999999999999\\
190	0.97999999999999\\
200	0.97999999999999\\
};
\addlegendentry{Proposed}
\end{axis}
\end{tikzpicture}%
\caption{\label{fig:optdim_acc2}Accuracy, $\frac{\textrm{Number of times }k^*=10}{\textrm{Number of trials}}$}
\end{subfigure}\hfill
\begin{subfigure}[t]{0.5\linewidth}
\centering
%% Figure 8b
\definecolor{mycolor1}{rgb}{0.98431,0.50196,0.44706}%
\definecolor{mycolor2}{rgb}{0.59608,0.30588,0.63922}%
\definecolor{mycolor3}{rgb}{0.99216,0.70588,0.38431}%
\definecolor{mycolor4}{rgb}{0.55294,0.82745,0.78039}%

\begin{tikzpicture}
\begin{axis}[%
width=.95\linewidth,
at={(0.614in,0.52in)},
xmin=20,
xmax=200,
xlabel style={font=\color{white!15!black}},
xlabel={Ambient dimension, $n$},
xtick={20,50,100,150,200},
ymin=7,
ymax=20,
ylabel style={font=\color{white!15!black}},
ylabel={Mean selected order},
axis background/.style={fill=white},
grid=both,
grid style={line width=.1pt, draw=gray!10},
major grid style={line width=.2pt,draw=gray!50},
legend style={legend cell align=left, align=left, draw=white!15!black}
]
% optimal dimension line
\addplot [color=black, line width=1.0pt]
  table[row sep=crcr]{%
20	10\\
200	10\\
};

% SVD
\addplot [color=mycolor3, dashdotted, line width=2.0pt, mark=x, mark options={solid, mycolor3}]
  table[row sep=crcr]{%
20	8.78999999999999\\
30	14.22\\
40	15\\
50	15\\
60	15\\
70	15\\
80	15\\
90	15\\
100	15\\
110	15\\
120	15\\
130	15\\
140	15\\
150	15\\
160	15\\
170	15\\
180	15.09\\
190	15\\
200	15.16\\
};
%\addlegendentry{SVD}

% Hybrid
\addplot [color=mycolor4, densely dotted, line width=2.0pt, mark=square, mark options={solid, mycolor4}]
  table[row sep=crcr]{%
20	18.54\\
30	10.33\\
40	10.24\\
50	10.06\\
60	10.07\\
70	10.05\\
80	10\\
90	10.01\\
100	10\\
110	10.01\\
120	10\\
130	10\\
140	10\\
150	10\\
160	10.01\\
170	10\\
180	10\\
190	10\\
200	10\\
};
%\addlegendentry{GMEB + 29}

% Santamaria
\addplot [color=mycolor1, line width=2.0pt, mark=o, mark options={solid, mycolor1}]
  table[row sep=crcr]{%
20	18.59\\
30	15\\
40	15\\
50	14.97\\
60	14.73\\
70	14.14\\
80	13.41\\
90	12.62\\
100	11.63\\
110	11\\
120	10.59\\
130	10.24\\
140	10.1\\
150	10.04\\
160	10.02\\
170	10\\
180	10\\
190	10\\
200	10\\
};
%\addlegendentry{Santamaria et al.}

% Proposed
\addplot [color=mycolor2, dashed, mark=triangle, mark options={solid, mycolor2}, line width=2.0pt]
  table[row sep=crcr]{%
20	16.01\\
30	15.62\\
40	13.64\\
50	11.6\\
60	11\\
70	11.1\\
80	10.4\\
90	10.2\\
100	10.2\\
110	10.2\\
120	10.4\\
130	10.2\\
140	10.2\\
150	10.6\\
160	10.1\\
170	10.1\\
180	10.2\\
190	10.2\\
200	10.2\\
};
%\addlegendentry{Proposed}

% Optimal dimension label
\node[anchor=west] at (22,9.0){\small $k^* = 10$};
\end{axis}
\end{tikzpicture}%
\caption{\label{fig:optdim_mv2}Mean selected order}
\end{subfigure}
\caption{\label{fig:opt_dim2} Order-selection accuracy and mean selected order relative to the ambient dimension of the data from $100$ Monte Carlo trials using the proposed order-selection rule (purple dashed line with triangle markers), the method of Santamar{\'\i}a \etal~\cite{santamaria2016order}  (pink solid line with circle markers), the hybrid method (turquoise dotted line with square markers), and the elbow point of the SVD (orange dash-dotted line with circle markers).  The data consists $20$ points from $\coprod_{p \in \mathcal{P}}{\textrm{Gr}(p,n)}$ for $\mathcal{P} = \{10,11, \ldots,20\}$ and $n = 20, 30, \ldots, 200$ with an SNR of $9$ generated according to the model in Section~\ref{subsec:nested}.}
\end{figure*}

Figure~\ref{fig:optdim_acc2} shows the percentage of $100$ Monte Carlo trials for which the proposed order-selection rule (purple dashed line with triangle markers), the method of Santamar{\'\i}a \etal~\cite{santamaria2016order}  (pink solid line with circle markers), the hybrid method (turquoise dotted line with square markers), and the elbow point of the SVD (orange dash-dotted line with circle markers) were able to correctly identify the optimal order of the common subspace relative to the ambient dimension. Figure~\ref{fig:optdim_mv2} shows the mean selected order, averaged across all trials. We can see that when the ambient dimension is small, all methods other than the SVD tend to overestimate the order of the common subspace.  This is a result of the noise dimensions being relatively close in the low-dimensional spaces. The dimension of Gr$(k,n)$ is $k(n-k),$ so for $k\approx \max_i\{p_i\} \approx n$ all samples are very similar regardless of the data model. As the ambient dimension grows and the randomly selected dimensions become further apart on average, the proposed method and the hybrid method correctly select the order with a high degree of accuracy.  The proposed method achieves slightly lower accuracy and has less stable performance than the hybrid method because $c_{\textrm{pen}}(k)$ can be significantly affected by even one subspace that is similar to $\*U^{*\perp}(k)$. However, this behavior is consistent with the assumption that every sample is valid and there are no outliers in the collection of data. As expected, \cite{santamaria2016order} initially estimates the order as the dimension of the common subspace for the smaller ball and over-estimates the order as $15$, while the two methods that rely on the minimum enclosing ball estimate the dimension of the common subspace for that support set. Predictably, the elbow point of the SVD has a very low accuracy regardless of the ambient dimension.  In essence, this method is attempting to preserve all dimensions that are not pure noise.

\begin{figure*}[!t]
\begin{subfigure}[t]{0.5\linewidth}
\centering
%% Figure 9a
\definecolor{mycolor1}{rgb}{0.98431,0.50196,0.44706}%
\definecolor{mycolor2}{rgb}{0.59608,0.30588,0.63922}%
\definecolor{mycolor3}{rgb}{0.99216,0.70588,0.38431}%
\definecolor{mycolor4}{rgb}{0.55294,0.82745,0.78039}%

\begin{tikzpicture}
\begin{axis}[%
width=.95\linewidth,
at={(0.614in,0.52in)},
xmin=-5,
xmax=10,
xlabel style={font=\color{white!15!black}},
xlabel={SNR (dB)},
ymin=0,
ymax=1,
ylabel style={font=\color{white!15!black}},
ylabel={Accuracy},
grid=both,
grid style={line width=.1pt, draw=gray!10},
major grid style={line width=.2pt,draw=gray!50},
axis background/.style={fill=white},
reverse legend,
legend style={legend cell align=left, align=left, at={(0.97,0.3)},anchor=east, draw=white!15!black}
]

% SVD
\addplot [color=mycolor3, dashdotted, line width=2.0pt, mark=x, mark options={solid, mycolor3}]
  table[row sep=crcr]{%
-5	0\\
-4	0\\
-3	0\\
-2	0\\
-1	0\\
0	0\\
1 0\\
2	0\\
3	0\\
4	0\\
5	0\\
6	0\\
7	0\\
8	0\\
9	0\\
10	0\\
};
\addlegendentry{SVD}

% Hybrid
\addplot [color=mycolor4, densely dotted, line width=2.0pt, mark=square, mark options={solid, mycolor4}]
  table[row sep=crcr]{%
-5 0\\
-4	0\\
-3	0\\
-2	0\\
-10\\
0	0\\
1	0\\
2	0.790000000000001\\
3	1\\
4	1\\
5 1\\
6	1\\
7	1\\
8	1\\
9	1\\
10	1\\
};
\addlegendentry{Hybrid}

% Santamaria
\addplot [color=mycolor1, line width=2.0pt, mark=o, mark options={solid, mycolor1}]
  table[row sep=crcr]{%
-5	0\\
-4	0\\
-3	0\\
-2	0\\
-1	0\\
0	1\\
1	1\\
2	1\\
3	1\\
4	1\\
5	1\\
6	1\\
71\\
8	1\\
9	1\\
10	1\\
};
\addlegendentry{\cite{santamaria2016order}}

% Proposed
\addplot [color=mycolor2, dashed, mark=triangle, mark options={solid, mycolor2}, line width=2.0pt]
  table[row sep=crcr]{%
-5	0\\
-4	0\\
-3	0\\
-2	0\\
-1	0\\
0	0\\
1	0.27\\
2	1\\
3	1\\
4	1\\
5	1\\
6	1\\
7	1\\
8	1\\
9	1\\
10	1\\
};
\addlegendentry{Proposed}
\end{axis}
\end{tikzpicture}%
\caption{\label{fig:optdim_acc1}Accuracy, $\frac{\textrm{Number of times }k^*=3}{\textrm{Number of trials}}$}
\end{subfigure}\hfill
\begin{subfigure}[t]{0.5\linewidth}
\centering
%% Figure 9b
\definecolor{mycolor1}{rgb}{0.98431,0.50196,0.44706}%
\definecolor{mycolor2}{rgb}{0.59608,0.30588,0.63922}%
\definecolor{mycolor3}{rgb}{0.99216,0.70588,0.38431}%
\definecolor{mycolor4}{rgb}{0.55294,0.82745,0.78039}%

\begin{tikzpicture}
\begin{axis}[%
width=.95\linewidth,
at={(0.614in,0.52in)},
xmin=-5,
xmax=10,
xlabel style={font=\color{white!15!black}},
xlabel={SNR (dB)},
ymin=0,
ymax=5,
ytick={1,2,3,4,5},
ylabel style={font=\color{white!15!black}},
ylabel={Mean selected order},
grid=both,
grid style={line width=.1pt, draw=gray!10},
major grid style={line width=.2pt,draw=gray!50},
axis background/.style={fill=white},
legend style={legend cell align=left, align=left, draw=white!15!black}
]

% Optimal dimension line
\addplot [color=black, line width=1.0pt, forget plot]
  table[row sep=crcr]{%
10	3\\
-5	3\\
};
%\addlegendentry{$\text{k}^\text{*}\text{ = 3}$}

% SVD
\addplot [color=mycolor3, dashdotted, line width=2.0pt, mark=x, mark options={solid, mycolor3}]
  table[row sep=crcr]{%
-5	5\\
-4	5\\
-3	5\\
-2 5\\
-1	5\\
0	5\\
1	5\\
2	5\\
3	5\\
4	5\\
5	5\\
6	5\\
7	5\\
8	5\\
9	5\\
10 5\\
};
%\addlegendentry{SVD}

% Hybrid
\addplot [color=mycolor4, densely dotted, line width=2.0pt, mark=square, mark options={solid, mycolor4}]
  table[row sep=crcr]{%
-5	0\\
-4	0\\
-3	0\\
-2	0\\
-1	0\\
0	0\\
1	0\\
2	2.75\\
3	3\\
4	3\\
5	3\\
6	3\\
7	3\\
8	3\\
9	3\\
10	3\\
};
%\addlegendentry{GMEB + 29}

% Santamaria
\addplot [color=mycolor1, line width=2.0pt, mark=o, mark options={solid, mycolor1}]
  table[row sep=crcr]{%
-5	0\\
-4	0\\
-3	0\\
-2	0\\
-1	0\\
0	3\\
1	3\\
2	3\\
3 3\\
4	3\\
5	3\\
6	3\\
7 3\\
8	3\\
9	3\\
10	3\\
};
%\addlegendentry{Santamaria et al.}

% Proposed
\addplot [color=mycolor2, dashed, mark=triangle, mark options={solid, mycolor2}, line width=2.0pt]
  table[row sep=crcr]{%
-5	0\\
-4	0\\
-3	0\\
-2	0\\
-1	0\\
0	0\\
1	0.81\\
2 3\\
3	3\\
4	3\\
5	3\\
6	3\\
7	3\\
8	3\\
9	3\\
10	3\\
};
%\addlegendentry{Proposed}

% Optimal dimension label
\node[] at (-3,3.25){\small $k^* = 3$};
\end{axis}
\end{tikzpicture}%
\caption{\label{fig:optdim_mv1}Mean selected order}
\end{subfigure}
\caption{\label{fig:opt_dim1} Order-selection accuracy and mean selected order relative to the signal-to-noise ratio of the data (in dB) from $100$ Monte Carlo trials using the proposed order-selection rule (purple dashed line with triangle markers), the method of Santamar{\'\i}a \etal~\cite{santamaria2016order}  (pink solid line with circle markers), the hybrid method (turquoise dotted line with square markers), and the elbow point of the SVD (orange dash-dotted line with circle markers). The data consists $225$ points from $\coprod_{p \in \mathcal{P}}{\textrm{Gr}(p,100)}$ for $\mathcal{P} = \{3,4,5\}$ generated according to the model in Section~\ref{subsec:nonuni}.}
\end{figure*}
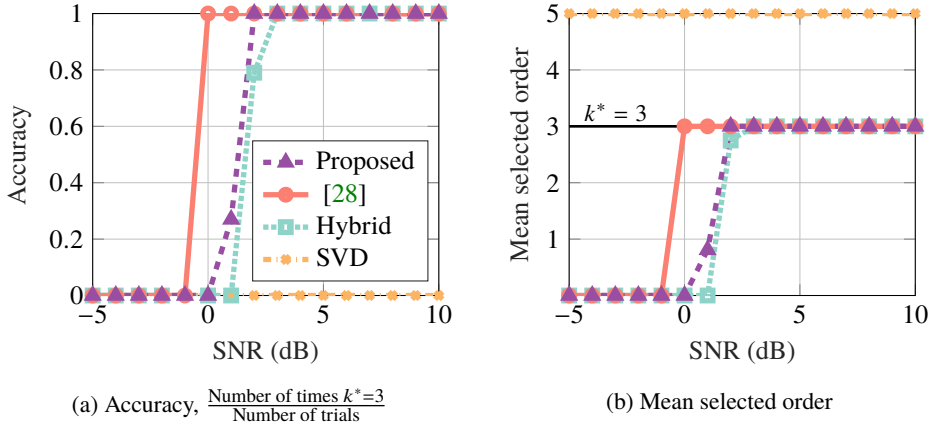

Figure~\ref{fig:opt_dim1} shows a comparison using data from the second model, a ball that is sampled more densely from a random arc. For some $\*Z_1 \in \textrm{Gr}(3,100),$ $M_1=200$ points are sampled uniformly from $\mathcal{B}_{0.5}(\*Z_1) \subset \textrm{Gr}(3,100)$ and $M_2 = 25$ additional points are then sampled from a random arc on the same ball. No points were sampled from the interior of the ball. Each of these $M = 225$ subspaces is completed to basis for a point on Gr$(p_i,100)$ for $p_i \in \{3,4,5\},$ and zero-mean Gaussian noise is added to each basis to create noisy data sets. In this experiment, the ambient dimension is fixed and we allow the SNR to vary from $-5$dB to $10$dB.

With this data the optimal order of the common subspace is $k^{*} = 3$ and center of the ball is $\*U^{*}(3) = \*Z_1.$ Figure~\ref{fig:optdim_acc1} shows the percentage of $100$ Monte Carlo trials for which the proposed order-selection rule (purple dashed line with triangle markers), the method of Santamar{\'\i}a \etal~\cite{santamaria2016order}  (pink solid line with circle markers), the hybrid method (turquoise dotted line with square markers), and the elbow point of the SVD (orange dash-dotted line with circle markers) were able to correctly identify the optimal order of the common subspace relative to the signal-to-noise ratio. Figure~\ref{fig:optdim_mv1} shows the mean selected order in the same trials. This experiment demonstrates the behavior of the different rules when all of the points are in the support of the minimum enclosing ball on Gr$(k^*,n)$. Each of the subspace averaging methods should theoretically select the same order in this experiment, because all of the points share the same number of dimensions and there is no ambiguity about the optimal solution.  Thus even though the mean computed by~\cite{santamaria2016order} is not the same point as the center of the GMEB, they lead to the same estimated rank. We see that in this scenario, the behavior of the rules using $\ell_{\infty}$-norm and the $\ell_2$-norm are similar with a sharp phase transition when the power of the signal and the power of the noise are almost equal, although the $\ell_2$-norm transitions to the correct order at a slightly higher noise power. This suggests that for situations where the data is free from outliers and the $\ell_{\infty}$-mean is close to the $\ell_2$-mean, either technique will accurately estimate the number of common dimensions. The elbow point of the singular value decomposition fails to identify the common dimension in all trials.

\begin{figure*}[!t]
\begin{subfigure}[t]{0.5\linewidth}
\centering
% Figure 10a
\definecolor{mycolor1}{rgb}{0.98431,0.50196,0.44706}%
\definecolor{mycolor2}{rgb}{0.59608,0.30588,0.63922}%
\definecolor{mycolor3}{rgb}{0.99216,0.70588,0.38431}%
\definecolor{mycolor4}{rgb}{0.55294,0.82745,0.78039}%

\begin{tikzpicture}
\begin{axis}[%
width=.95\linewidth,
at={(0.614in,0.52in)},
xmin=5,
xmax=40,
xlabel style={font=\color{white!15!black}},
xlabel={Ambient dimension, $n$},
xtick={5,10,15,20,25,30,35,40},
ymin=0,
ymax=1,
ylabel style={font=\color{white!15!black}},
ylabel={Accuracy},
axis background/.style={fill=white},
reverse legend,
grid=both,
grid style={line width=.1pt, draw=gray!10},
major grid style={line width=.2pt,draw=gray!50},
legend style={legend cell align=left, align=left, at={(0.97,0.3)},anchor=east, draw=white!15!black}
]

% SVD
\addplot [color=mycolor3, dashdotted, line width=2.0pt, mark=x, mark options={solid, mycolor3}]
  table[row sep=crcr]{%
5	0\\
6	0\\
7	0\\
8	0\\
9	0\\
10	0\\
11	0\\
12	0\\
13	0\\
14	0\\
15	0\\
20	0\\
25	0\\
30	0\\
35	0\\
40	0\\
};
\addlegendentry{SVD}

% Hybrid
\addplot [color=mycolor4, densely dotted, line width=2.0pt, mark=square, mark options={solid, mycolor4}]
  table[row sep=crcr]{%
5	0\\
6	0\\
7	0\\
8	0.530000000000001\\
9	0.909999999999997\\
10	0.990000000000002\\
11	1\\
12	1\\
13	1\\
14	1\\
15	1\\
20	1\\
25	1\\
30	1\\
35	1\\
40	1\\
};
\addlegendentry{Hybrid}

% Santamaria
\addplot [color=mycolor1, line width=2.0pt, mark=o, mark options={solid, mycolor1}]
  table[row sep=crcr]{%
5	0\\
6	0\\
7	0\\
8	0\\
9	0.00999999999999801\\
10	0.299999999999997\\
11	0.93\\
12	1\\
13	1\\
14	1\\
15	1\\
20	1\\
25	1\\
30	1\\
35	1\\
40	1\\
};
\addlegendentry{\cite{santamaria2016order}}

% Proposed
\addplot [color=mycolor2, dashed, mark=triangle, mark options={solid, mycolor2}, line width=2.0pt]
  table[row sep=crcr]{%
5	0\\
6	0\\
7	0\\
8	0\\
9	0\\
10	0\\
11	0\\
12	0.0799999999999983\\
13	0.270000000000003\\
14	0.840000000000003\\
15	0.93\\
20	1\\
25	1\\
30	1\\
35	1\\
40	1\\
};
\addlegendentry{Proposed}
\end{axis}
\end{tikzpicture}%
\caption{\label{fig:optdim_acc3}Accuracy, $\frac{\textrm{Number of times }k^*=0}{\textrm{Number of trials}}$}
\end{subfigure}\hfill
\begin{subfigure}[t]{0.5\linewidth}
\centering
%% Figure 10b
\definecolor{mycolor1}{rgb}{0.98431,0.50196,0.44706}%
\definecolor{mycolor2}{rgb}{0.59608,0.30588,0.63922}%
\definecolor{mycolor3}{rgb}{0.99216,0.70588,0.38431}%
\definecolor{mycolor4}{rgb}{0.55294,0.82745,0.78039}%

\begin{tikzpicture}
\begin{axis}[%
width=.95\linewidth,
at={(0.614in,0.52in)},
xmin=5,
xmax=40,
xlabel style={font=\color{white!15!black}},
xlabel={Ambient dimension, $n$},
xtick={5,10,15,20,25,30,35,40},
ymin=-2,
ymax=14,
ylabel style={font=\color{white!15!black}},
ytick={0,2,4,6,8,10,12,14},
ylabel={Mean selected order},
axis background/.style={fill=white},
grid=both,
grid style={line width=.1pt, draw=gray!10},
major grid style={line width=.2pt,draw=gray!50},
legend style={legend cell align=left, align=left, draw=white!15!black}
]

% Optimal dimension line
\addplot [color=black, line width=1.0pt]
  table[row sep=crcr]{%
20	0\\
200	0\\
};
%\addlegendentry{$\text{k}^\text{*}\text{ = 3}$}

% SVD
\addplot [color=mycolor3, dashed, line width=2.0pt, mark=x, mark options={solid, mycolor3}]
  table[row sep=crcr]{%
5	2.12\\
6	2.66\\
7	3.38\\
8	3.96\\
9	4.75\\
10	5.46\\
11	5.52\\
12	6.37\\
13	5.94\\
14	7.16\\
15	7.66\\
20	11.38\\
25	12.7\\
30	10.86\\
35	13.94\\
40	13.16\\
};
%\addlegendentry{SVD}

% Hybrid
\addplot [color=mycolor4, dashed, line width=2.0pt, mark=square, mark options={solid, mycolor4}]
  table[row sep=crcr]{%
5	5\\
6	5\\
7	5\\
8	1.53\\
9	0.210000000000001\\
10	0.0200000000000031\\
11	0\\
12	0\\
13	0\\
14	0\\
15	0\\
20	0\\
25	0\\
30	0\\
35	0\\
40	0\\
};
%\addlegendentry{GMEB + 29}

% Santamaria
\addplot [color=mycolor1, line width=2.0pt, mark=o, mark options={solid, mycolor1}]
  table[row sep=crcr]{%
5	5\\
6	6\\
7	5.79\\
8	4.08\\
9	2.13\\
10	0.880000000000003\\
11	0.0700000000000003\\
12	0\\
13	0\\
14	0\\
15	0\\
20	0\\
25	0\\
30	0\\
35	0\\
40	0\\
};
%\addlegendentry{Santamaria et al.}

% Proposed
\addplot [color=mycolor2, dashed, mark=triangle, mark options={solid, mycolor2}, line width=2.0pt]
  table[row sep=crcr]{%
5	4\\
6	5\\
7	4.99\\
8	5\\
9	5\\
10	4.99\\
11	5\\
12	4.6\\
13	3.65\\
14	0.799999999999997\\
15	0.350000000000001\\
20	0\\
25	0\\
30	0\\
35	0\\
40	0\\
};
%\addlegendentry{Proposed}

% Optimal dimension label
\node[anchor=west] at (22,-1.0){\small $k^* = 0$};
\end{axis}
\end{tikzpicture}%
\caption{\label{fig:optdim_mv3}Mean selected order}
\end{subfigure}
\caption{\label{fig:opt_dim3} Order-selection accuracy and mean selected order relative to the ambient dimension of the data when there is no common subspace. Results are from $100$ Monte Carlo trials using the proposed order-selection rule (purple dashed line with triangle markers), the method of Santamar{\'\i}a \etal~\cite{santamaria2016order}  (pink solid line with circle markers), the hybrid method (turquoise dotted line with square markers), and the elbow point of the SVD (orange dash-dotted line with circle markers).  The data consists $50$ points from $\coprod_{p \in \mathcal{P}}{\textrm{Gr}(p,n)}$ for $\mathcal{P} = \{3,4,5\}$ and $n = 5, 6, \ldots, 15,20,25,\ldots, 40.$}
\end{figure*}
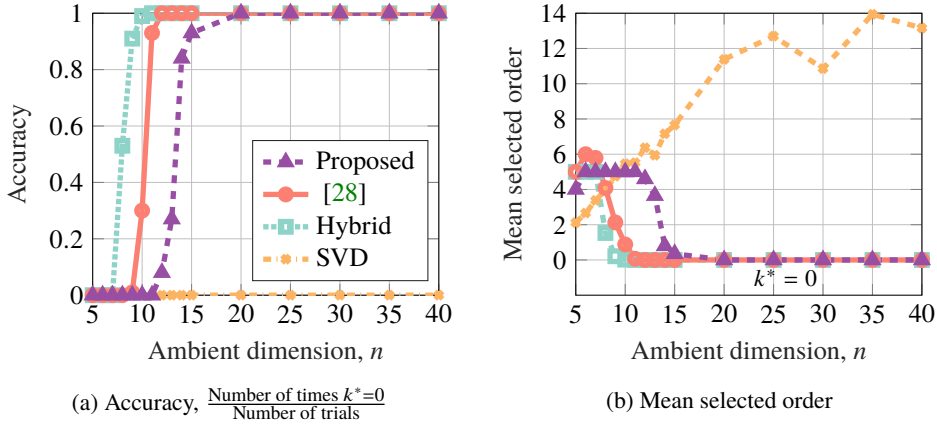

Finally, in Figure~\ref{fig:opt_dim3} we see the ability of each method to identify when there is no subspace common to a collection of points. This is a valuable test because estimating $k^*=0$ suggests that there is no information shared across all the data and that averaging the points is not an appropriate way to aggregate the information in the data. The data in this experiment consists of $50$ subspaces chosen uniformly at random from Gr$(p_i,n)$ for $p_i \in \{3,4,5\}$ for $i = 1, \ldots, 10$ with ambient dimensions $n = 5, 6, \ldots, 15,20,25,\ldots, 40.$ The noise variance does not affect performance in this task because there is no signal so SNR undefined. In Figure~\ref{fig:optdim_acc3} we see a similar phase transition to that of Figure~\ref{fig:opt_dim1}. The hybrid method is able to achieve perfect accuracy for ambient dimensions greater than $10,$ while \cite{santamaria2016order} and the proposed method transition shortly thereafter. The SVD fails every time, but that is to be expected in this scenario. The elbow point method computes two lines that minimize the residual for the scree plot, and chooses dimension as the index of the singular value just larger than the intersection of those lines.  A line cannot be fit to zero points, so the method will not select $k^*=0$ or $k^* = n$ as a solution. However, in Figure~\ref{fig:optdim_mv3} we see that the SVD is significantly overestimating the dimension of the (non-existent) common subspace, so the poor performance is not an issue of the method being unable to select $0$ as the optimal dimension. When $n$ is small the proposed algorithm incorrectly identifies a relationship between the subspaces, but as the ambient dimension grows the optimal order, $k^*=0$, is selected with increasing accuracy.  As noted in discussion of Figure~\ref{fig:opt_dim2}, the misidentifications in low dimensions are due to the minimum similarity between the points and $\*U^{*\perp}(k)$ being higher when $k\approx \max_i\{p_i\} \approx n$.

%% Section 8
\section{Conclusions}
\label{sec:conclusions}
The recent trend of performing machine learning tasks on linear subspace data has created a need for flexible subspace averages, ones that can be computed accurately and in a principled manner for subspaces of differing dimension. In response to this need, we have proposed an algorithm to find the $\ell_{\infty}$-center of mass using a subgradient algorithm to solve the dual problem with respect to a point-to-set distance.  We additionally proposed a flexible data generation model to create subspaces of differing dimensions with ground-truth for the GMEB that emulates realistic settings where an $\ell_{\infty}$-average would be appropriate.  On this synthetic data, the proposed algorithm provides estimates of the GMEB center with high accuracy. However, the high computational complexity means that an existing primal method can provide low-accuracy solutions more quickly for large data sets. One direction for future expansion is to develop a core-set theory akin to that of~\cite{badoiu2003smaller} in order to estimate the GMEB on a subset of the data with theoretical accuracy guarantees. A related area for further study is to develop an active-set approach for $\ell_{\infty}$-averaging of mixed-dimensional subspaces, \`{a} la John~\cite{john2014extremum}. Active-set methods also attempt to minimize the cost function over a subset of the data. However, the active-set approach looks for a subset of the data that solves the original problem exactly, whereas the core-set technique computes error bounds on the solution provided by \textit{any} subset of a given size.  One theoretical hurdle to achieving an active-set method is a theorem on the minimum number of points required to define a Grassmannian ball given a fixed Grassmann manifold and subspaces of differing dimensions. 

Finally, we proposed a geometric order-fitting rule that estimates the best dimension for the common subspace. This rule fits the common dimensions of the subspaces in the support set of the minimum enclosing ball, which is appropriate for data where all subspace samples are assumed to be valid examples of the model of interest. We additionally implement a hybrid technique for estimating the dimension of the common subspace that modifies the order-selection rule of~\cite{santamaria2016order} for use with the $\ell_{\infty}$-average. This hybrid method would not be possible for existing techniques that estimate the GMEB, because it uses the values of the dual variables as weights for an eigenvalue decomposition at each potential order.  The hybrid approach outperforms the proposed technique and that of~\cite{santamaria2016order} when the ambient dimension is close to the subspace dimension of the data points.

A high-accuracy estimate of the GMEB center combined with an order-selection rule for the number of common dimensions results in a powerful technique for detecting and estimating similarity in a collection of subspaces. We anticipate that many practical applications will arise in the form of distributed large-scale problems, where the subspace averaging can be used for aggregation, for example the sparse subspace clustering of~\cite{abdolali2019scalable}. 

%% Endmatter

\section*{Acknowledgments}
The authors would like to thank Emilie Renard for the stimulating discussions that improved the ideas presented here.

\bibliographystyle{siamplain}
\bibliography{gmeb_refs}

\newpage
\appendix
%% Appendix A: Algorithm
\section{GMEB dual subgradient algorithm}
\label{sec:alg}
\begin{algorithm*}[!ht] 
	\caption{Algorithm to minimize Equation~\eqref{eq:dualProb} with back-tracking line search}
	\label{alg:subgrad}
	\begin{spacing}{1.25}
	\begin{algorithmic}[1]
	  \Function{GMEB}{$\big\{\mathbf{X}_i\big\}_{i=1}^M,k,a,\eta,\zeta, \beta$}
		\State \textbf{input:} Data: $\big\{\mathbf{X}_i\big\}_{i=1}^M$, Rank: $k$, Step size parameter: $a$, Stopping criteria: $\eta$,  Step size threshold: $\zeta$, Growth parameter: $\beta$
 	  \State \textbf{output:} Weights: $\bm{\lambda}^*$, Minimax center: $\*U^*$
		\State $t \gets		 0$
	  \State $\bm{\lambda}^{(t)} \gets [\nicefrac{1}{M}, \ldots , \nicefrac{1}{M}]^T \in \mathbb{R}^M$ \Comment{$\bm{\lambda}^{(t)} \gets \bm{\lambda}^*(k-1)$ for warm-start}
		\State  $\*U^{(t)} \gets \textrm{dominant }k\textrm{ eigenvectors}\big(\sum_{i=1}^M \lambda_i^{(t)} X_i^{} X_i^T\big)$
		\State $\*g^{(t)} \gets -\big[d_{\textrm{Gr}(k,n)}(\*U^{(t)},\*X_1), d_{\textrm{Gr}(k,n)}(\*U^{(t)},\*X_2), \ldots, d_{\textrm{Gr}(k,n)}(\*U^{(t)},\*X_M) \big]^T$
		\State $f_{\textrm{primal}}(\*U^{(t)}) \gets \min_{i=1,\ldots,M} \{-d_{\textrm{Gr}(k,n)}(\*U^{(t)},\*X_i)\}$ \Comment{Primal cost at iteration $t$}
		\State $f_{\textrm{dual}}(\bm{\lambda}^{(t)}) \gets \bm{\lambda}^{(t)T}\*g^{(t)}$	 \Comment{Dual cost at iteration $t$}
%% Main loop
		\While{ $f_{\textrm{dual}}(\bm{\lambda}^{(t)})  - f_{\textrm{primal}}(\*U^{(t)}) > \eta$ \AND $\underset{i=1,\ldots,10}{\max}\{f_{\textrm{dual}}(\bm{\lambda}^{(t-i)})-f_{\textrm{dual}}(\bm{\lambda}^{(t)})\}> \eta$}		
			\State $t \gets t + 1$
			\State $\alpha^{(t)} \gets \nicefrac{a}{\sqrt{t}}$
			\State $\bm{\lambda}^{(t)} \gets \bm{\lambda}^{(t-1)} - \alpha^{(t)} \*g^{(t-1)}$, $\bm{\lambda}^{(t)} \gets \nicefrac{\bm{\lambda}^{(t)}}{\|\bm{\lambda}^{(t)}\|_1}$
			\State  $\*U^{(t)} \gets \textrm{dominant }k\textrm{ eigenvectors}\big(\sum_{i=1}^M \lambda_i^{(t)} \*X_i^{} \*X_i^T\big)$
			\State $\*g^{(t)} \gets  -\big[d_{\textrm{Gr}(k,n)}(\*U^{(t)},\*X_1), d_{\textrm{Gr}(k,n)}(\*U^{(t)},\*X_2), \ldots, d_{\textrm{Gr}(k,n)}(\*U^{(t)},\*X_M) \big]^T$
			\State $\tilde{\alpha}^{(t)} \gets \alpha^{(t)}$
			\State $\tilde{\bm{\lambda}}^{(t)} \gets \bm{\lambda}^{(t)}$	
			\State $f_{\textrm{dual}}(\tilde{\bm{\lambda}}^{(t)}) \gets \tilde{\bm{\lambda}}^{(t)T}\*g^{(t)}$
%% Line search loop
			\While{ $f_{\textrm{dual}}( \tilde{\bm{\lambda}}^{(t)} ) > f_{\textrm{dual}}( \bm{\lambda}^{(t-1)}) \AND \tilde{\alpha}^{(t)} > \zeta\alpha^{(t)}$} \Comment{Back-tracking line search} 
				\State $a \gets \nicefrac{a}{2}$
				\State $\tilde{\alpha}^{(t)}\gets \nicefrac{a}{\sqrt{t}}$
				\State $\tilde{\bm{\lambda}}^{(t)} \gets \bm{\lambda}^{(t-1)} - \tilde{\alpha}^{(t)} \*g^{(t-1)}$, $\tilde{\bm{\lambda}}^{(t)} \gets \nicefrac{\tilde{\bm{\lambda}}^{(t)}}{\|\tilde{\bm{\lambda}}^{(t)}\|_1}$
				\State  $\tilde{\*U}^{(t)} \gets \textrm{dominant }k\textrm{ eigenvectors}\big(\sum_{i=1}^M \tilde{\lambda}_i^{(t)} \*X_i^{} \*X_i^T\big)$
				\State $\tilde{\*g}^{(t)} \gets  -\big[ d_{\textrm{Gr}(k,n)}(\tilde{\*U}^{(t)},\*X_1), d_{\textrm{Gr}(k,n)}(\tilde{\*U}^{(t)},\*X_2), \ldots, d_{\textrm{Gr}(k,n)}(\tilde{\*U}^{(t)},\*X_M) \big]^T$
				\State $f_{\textrm{dual}}(\tilde{\bm{\lambda}}^{(t)}) \gets \tilde{\bm{\lambda}}^{(t)T}\tilde{\*g}^{(t)}$
%% Update variables if cost decreases
				\If{ $f_{\textrm{dual}}( \tilde{\bm{\lambda}}^{(t)} ) \leq f_{\textrm{dual}}( \bm{\lambda}^{(t-1)})$} \Comment{Update variables if $f_\textrm{dual}$ decreases}
				\State $a \gets \beta a$
				\State $\bm{\lambda}^{(t)} \gets \tilde{\bm{\lambda}}^{(t)}$
				\State $\*U^{(t)} \gets \tilde{\*U}^{(t)}$
				\State $\*g^{(t)} \gets \tilde{\*g}^{(t)}$
				\EndIf
			\EndWhile
		\State $f_{\textrm{primal}}(\*U^{(t)}) \gets \min_{i=1,\ldots,M} \{-d_{\textrm{Gr}(k,n)}(\*U^{(t)},\*X_i)\}$
			\State $f_{\textrm{dual}}(\bm{\lambda}^{(t)}) \gets \bm{\lambda}^{(t)T}\*g^{(t)}$	
		\EndWhile
		
		\Return  $\bm{\lambda^{(t)}}, \ \*U^{(t)}$
		\EndFunction
	\end{algorithmic}
	\end{spacing}
\end{algorithm*}

\end{document}